\documentclass[10pt,journal,compsoc]{IEEEtran}
\usepackage{amsmath,amsfonts}
\usepackage{algorithmic}
\usepackage{array}
\usepackage[caption=false,font=normalsize,labelfont=sf,textfont=sf]{subfig}
\usepackage{textcomp}
\usepackage{stfloats}
\usepackage{url}
\usepackage{verbatim}
\usepackage{graphicx}
\usepackage{booktabs}
\usepackage{float}
\usepackage{subcaption}
\usepackage{cite}
\usepackage{amsmath}
\usepackage{cite}
\usepackage[colorlinks]{hyperref}
\usepackage[ruled,linesnumbered]{algorithm2e}
\usepackage{animate}
\usepackage{ragged2e}
\usepackage{multirow}
\usepackage{nth}
\usepackage[table]{xcolor}
\usepackage{makecell}
 
\usepackage{multicol}

\usepackage{balance}
\begin{document}
\title{StreamHOI: Interaction-aware Temporal Memory Adaptation for Streaming HOI Video Generation
}
\author{Zejing Rao, Haoxian Zhang, Xiaoqiang Liu, Yiping Meng, Guoxin Zhang, Pengfei Wan, Fan Tang, Tong-Yee Lee,~\IEEEmembership{Senior Member,~IEEE}%
\IEEEcompsocitemizethanks{
\IEEEcompsocthanksitem This work was conducted during the author's internship at Kling AI Research.
\IEEEcompsocthanksitem Corresponding author: Fan Tang, tfan.108@gmail.com
\IEEEcompsocthanksitem Z. Rao and F. Tang are with University of Chinese Academy of Sciences, Beijing, China. 
\IEEEcompsocthanksitem  H. Zhang, X. Liu, Y. Meng, G. Zhang and P. Wan are with Kling AI Research, Beijing, China.
\IEEEcompsocthanksitem T.-Y. Lee is with National Cheng Kung University, Tainan, Taiwan.
}
}

\IEEEtitleabstractindextext{%
\begin{abstract}
\justifying
Existing human--object interaction (HOI) video generation methods are largely limited to offline short-video generation with complex driving conditions, making them unsuitable for real-time interactive applications. 
We present \emph{StreamHOI}, a low-latency streaming framework for long-duration HOI video generation. 
Instead of converting heavily conditioned HOI pipelines into streaming systems, we study how an image-to-video streaming generator should organize historical memory to preserve interactions under bounded latency. 
We find that the standard sink-local memory design faces a trade-off in streaming HOI generation, and different transformer blocks show different historical-memory preferences for HOI regions and surrounding regions. 
To match memory composition with block behavior, StreamHOI performs offline HOI-aware block profiling and applies bias-guided memory-specialized training to adapt the generator to block-specific memory layouts. 
We further introduce a memory distance scaling module to strengthen long-range access to early interaction states. 
Extensive comparisons with both long-video baselines and recent HOI generation methods demonstrate that StreamHOI achieves strong interaction plausibility, object fidelity, human quality and efficiency, reaching 17.6 FPS with 0.75s first-chunk latency.
\end{abstract}


}
\maketitle

\section{Introduction}
\IEEEPARstart{H}{uman--object interaction} (HOI) has received increasing attention in computer vision and graphics~\cite{zuo2024graspdiff,hu2024hoimotion,xu2024anchorcrafter}, with applications in virtual humans, embodied agents, online object manipulation, and interactive content creation~\cite{xing2024make,zhang2025motioncrafter}. 
Existing HOI video generation methods~\cite{xu2024anchorcrafter,huang2025hyvideohoma,xue2024hoi,xu2026geohoi} improve controllability by conditioning on structured HOI cues. 
However, most of them follow an offline, fixed-length generation paradigm and mainly produce short clips. Such a paradigm fails to meet the needs of real-time virtual human interaction and online content creation, which require immediate response and continuous long-video generation.

\begin{figure}[tb]
  \centering
  \includegraphics[width=\columnwidth, alt={first comparison figure.}]{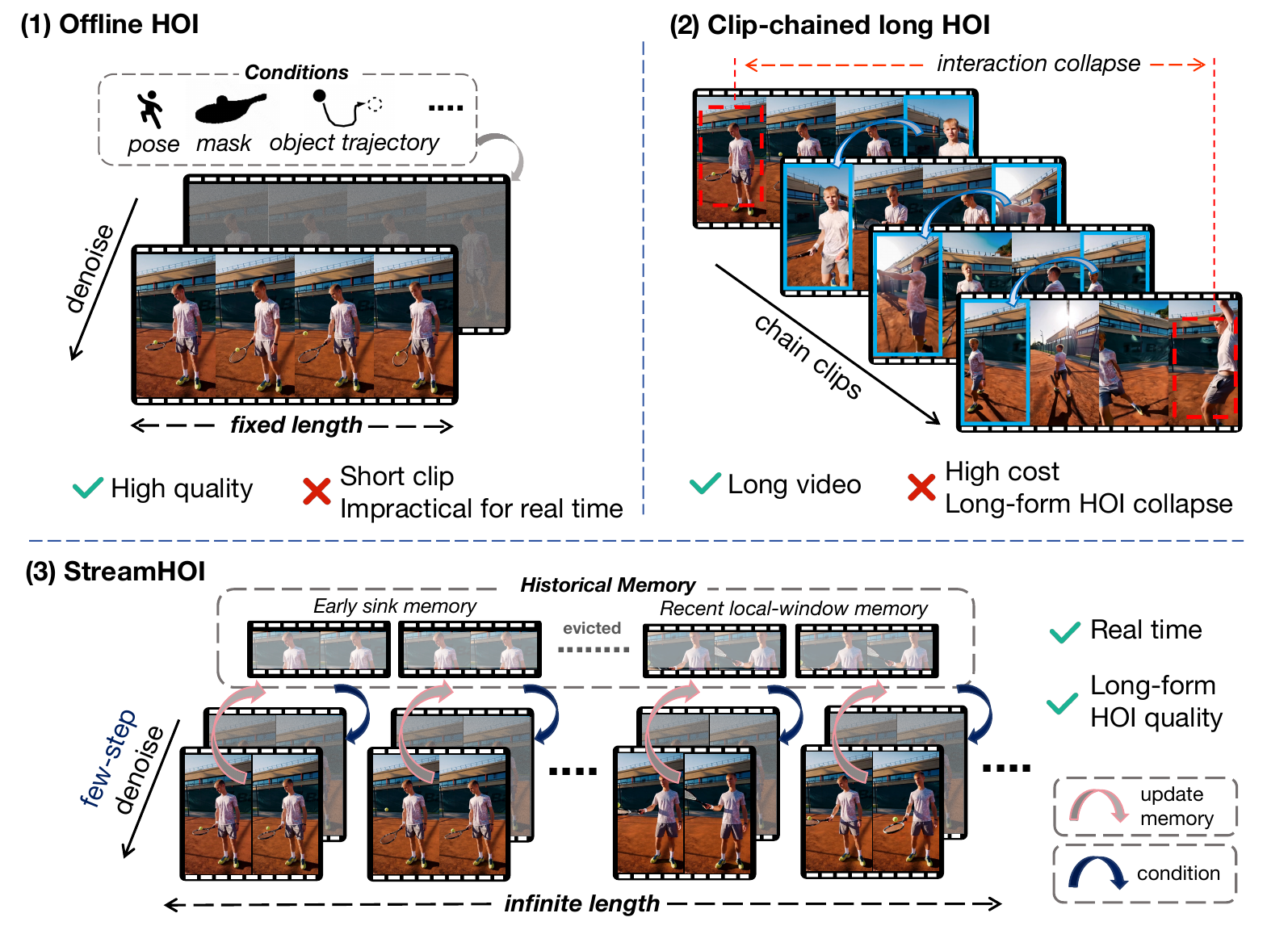}
  \caption{Conceptual comparison of HOI video generation paradigms. (1) Offline methods are restricted to fixed-length clips. (2) Clip-chained approaches extend duration but suffer from high computational cost and interaction collapse. (3) The proposed StreamHOI framework enables causal, low-latency streaming generation while maintaining robust long-horizon interaction consistency.}
  \label{fig:hoi-chain-ours}
\end{figure}

As illustrated in Fig.~\ref{fig:hoi-chain-ours}, a straightforward way to generate long HOI videos is to chain multiple short HOI clips, using the ending frames of one clip to initialize the next, as in InteractAvatar-long~\cite{zhang2026making}. 
This increases video duration, but still inherits the high offline generation cost of bidirectional HOI models and remains prone to interaction collapse in long-form generation. 
Streaming video generation~\cite{yin2025slow,huang2025selfforcing,zhu2026causal} provides a different paradigm by generating videos chunk by chunk with nearly constant computation. 
Such streaming generators rely on cached historical memory to condition each newly generated chunk. 
However, most short-video HOI methods depend on offline structured conditions, such as pose sequences, object trajectories, depth maps, or interaction layouts, which are difficult to obtain causally during streaming generation. 
Instead of converting heavily conditioned HOI methods into streaming pipelines, we focus on a more fundamental question: \textit{how should an image-to-video streaming generator organize historical memory to preserve human--object interactions under bounded latency?}

A common memory design for long streaming generation is to keep several early frames as sink memory together with a rolling local window~\cite{yang2025longlive}. 
Under a fixed memory budget, however, these two memory types create a trade-off in streaming HOI generation. 
Larger sink memory preserves early appearance and interaction states, improving subject and object consistency and hand--object relations, but it reduces recent context and may cause action jumps and motion discontinuity. 
Larger local memory improves short-term motion propagation, but it makes the model rely more on recent autoregressive frames and weakens long-term interaction preservation. 
In Sec.~\ref{sec:study}, we provide a detailed quantitative analysis of this trade-off.
We further observe that different Transformer blocks do not require the same memory composition: some blocks focus more on the subject, hands, object, and contact regions, while others focus more on surrounding scene and object-motion context. This motivates block-wise memory organization for streaming HOI generation.

In this study, we introduce \emph{StreamHOI}, a streaming HOI video generation framework that preserves long-duration human--object interactions through a simple yet effective block-wise organization of historical memory at the Transformer-block level. 
It first performs \emph{offline HOI-aware block profiling} to identify HOI-biased and surrounding-biased blocks according to their spatial historical-memory bias. Empirically, this profiling captures a stable block-wise memory tendency rather than sample-specific attention fluctuations.
To match memory composition with block behavior, we propose \emph{bias-guided memory-specialized training} (B-MST), which assigns block-specific sink-local memory compositions according to the profiled bias: HOI-biased blocks receive more sink memory for consistency-oriented use of reliable early interaction states, while surrounding-biased blocks receive more local-window memory for continuity-oriented use of recent motion and scene context around the human--object interaction. 
We further introduce a \emph{memory distance scaling} module, which shortens the perceived temporal distance between the current chunk and memory tokens in temporal RoPE, strengthening the access of HOI-biased blocks to distant interaction states. 

Our contributions are summarized as follows:
\begin{itemize}
  \item We identify a sink-local memory trade-off in streaming HOI generation and reveal block-wise historical-memory preferences through offline HOI-aware block profiling, showing that different transformer blocks exhibit HOI-biased and surrounding-biased behaviors.
  \item We propose bias-guided memory-specialized training, which trains the generator with block-specialized sink-local memory layouts and a memory distance scaling module, enabling different blocks to specialize in long-term interaction consistency and short-term motion continuity under a fixed memory budget.
  \item We validate StreamHOI through quantitative and qualitative comparisons, long-horizon evaluation, and ablation studies, demonstrating strong performance for streaming HOI video generation.
\end{itemize}

\section{Related Work}
\subsection{Human--Object Interaction Video Generation}
With the progress of controllable human video generation\cite{huang2024makeyouranchor,11359069,hu2025animateanyone2}, realistic human--object interaction (HOI) generation has become an important research direction. Human--object interaction video generation aims to synthesize videos in which humans and objects follow physically and semantically plausible relationships. Editing-based methods preserve interactions in real videos by replacing humans or objects, such as MIMO~\cite{men2025mimo} and AnimateAnyone2~\cite{hu2025animateanyone2} for human substitution, and HOI-Swap~\cite{xue2024hoi}, ReHoLD~\cite{fan2025rehold}, and iDiT-HOI~\cite{shen2025idithoi} for object replacement. Generation-based methods improve interaction controllability through explicit conditions such as human poses, object trajectories, depth maps, and bounding boxes. 
For example, SViMo~\cite{dang2025svimo} and HarmoHOI~\cite{dang2026harmohoiharmonizingappearance3d} explore video--motion co-generation for egocentric hand--object interactions. AnchorCrafter~\cite{xu2024anchorcrafter} uses multi-condition guidance to animate human--object interactions, DreamActor-H1~\cite{wang2025dreamactor-h1} retrieves motion templates and adjusts object layouts from reference images, and HunyuanVideo-HOMA~\cite{huang2025hyvideohoma} explores sparse user-editable conditions for controllable HOI generation. Recent works such as GeoHOI~\cite{xu2026geohoi} further improve object awareness and geometric interaction control, while InteractAvatar-long~\cite{zhang2026making} extends HOI generation to longer videos by chaining multiple generated clips.

However, existing HOI video generation methods are still mainly built on offline generation pipelines. Real-time generation and long-term interaction-state preservation in HOI video generation therefore remain underexplored. In contrast, \emph{StreamHOI} targets streaming HOI video generation, aiming to preserve human--object interaction states over long-duration generation while maintaining low-latency inference.

\subsection{Streaming and Long Video Generation}
Streaming video generation aims to convert offline video diffusion models into causal generators that produce videos chunk by chunk with low latency. Early causal video diffusion models are commonly trained with Teacher Forcing~\cite{jin2025pyramidal} or Diffusion Forcing~\cite{chen2024diffusion,song2025history}, where the model predicts future frames conditioned on clean or noisy history frames. To further reduce sampling cost, Consistency Distillation~\cite{song2023consistency,song2024improved} and Score Distillation~\cite{wang2023prolificdreamer,luo2023diff} have been used to distill multi-step diffusion models into few-step generators. A common instantiation of Score Distillation is Distribution Matching Distillation (DMD)~\cite{yin2024one}, which minimizes the KL divergence between the student and data distributions by following its gradient. CausVid~\cite{yin2025slow} distills a bidirectional video DiT into a causal student through ODE initialization and asymmetric DMD. Self-Forcing~\cite{huang2025selfforcing} reduces the train-test gap through self-rollout training, while Causal-Forcing~\cite{zhu2026causal} improves causal distillation by using an autoregressive teacher for causal ODE initialization.

Long video generation further requires preserving visual and semantic consistency beyond the short training horizon of video diffusion models. Recent methods improve long-horizon stability from different perspectives. FramePack~\cite{zhang2025framepack} compresses historical frames into a fixed-length context to reduce drift in next-frame prediction. LongLive~\cite{yang2025longlive}, Rolling Forcing~\cite{liu2025rolling}, and Self-Forcing++~\cite{cuiself} extend streaming generation through long-context training, self-rollout, rolling generation, or anti-drift strategies. Other methods treat historical context as memory. Context as Memory~\cite{yu2025cam}, Memory Forcing~\cite{huang2025memory}, MemFlow~\cite{ji2025memflow}, and Context Forcing~\cite{chen2026context} retrieve, compress, or organize historical memories to improve scene-level consistency. Another line of work modifies positional or memory mechanisms for longer extrapolation. Infinity-RoPE~\cite{yesiltepe2026infinity} reformulates temporal RoPE for infinite-horizon rollout, while Deep Forcing~\cite{yi2025deep} uses deep sink and participative compression to stabilize long videos without training. Recent heterogeneous KV methods, such as Head Forcing~\cite{tian2026headforcing}, further profile attention heads and assign head-specific cache strategies for long-range generation.

These methods are closely related to \emph{StreamHOI}, but differ in memory adaptation. Deep Forcing and Infinity-RoPE mainly rely on training-free, discrete RoPE recalibration or cache re-indexing, which may introduce boundary discontinuities when cached temporal segments are reassigned. In contrast, our MDS continuously reshapes temporal RoPE geometry within HOI-biased blocks. Heterogeneous KV methods such as Head Forcing use predefined head-level cache policies, whereas \emph{StreamHOI} learns interaction-aware block-level memory specialization to preserve fine-grained HOI states under bounded latency.
\begin{figure*}[tb]
  \centering
  \includegraphics[width=\textwidth, alt={Placeholder sink-local.}]{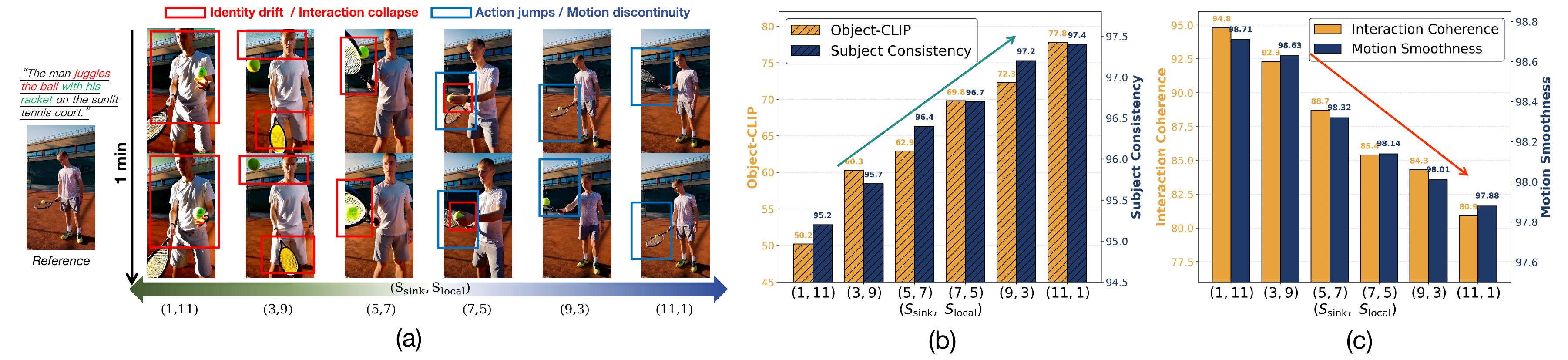}
  \caption{Illustration of the sink-local memory trade-off in streaming HOI generation under a fixed memory budget. (a) Qualitative impact of varying sink-to-local memory ratios, demonstrating long-term identity drift (insufficient sink memory) versus motion discontinuity (insufficient local memory). (b) Quantitative evaluation on object-clip and subject consistency. (c) Quantitative evaluation on interaction coherence and motion smoothness.}
  \label{fig:trade-off}
\end{figure*}

\begin{figure}[tb]
  \centering
  \includegraphics[width=\columnwidth, alt={Placeholder attention map figure.}]{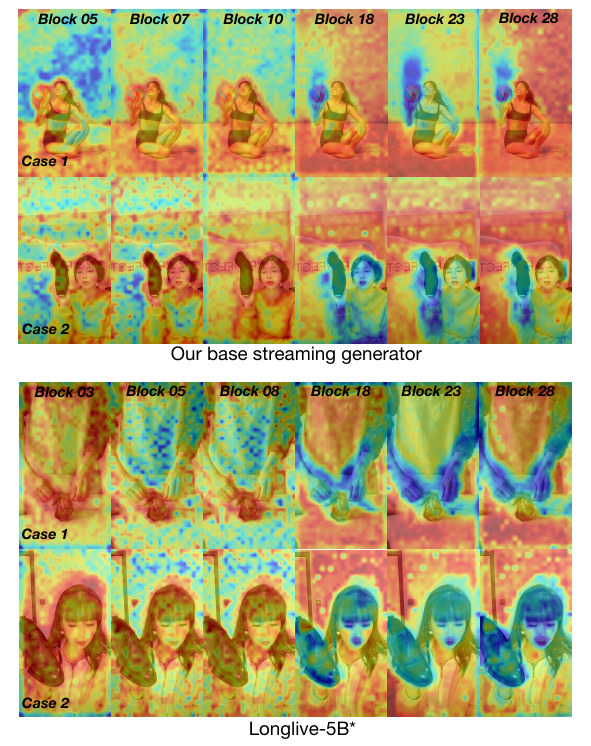}
  \caption{Block-wise historical-memory attention across two base streaming generators. Within each generator, spatial specialization is stable across diverse HOI examples; across generators, the exact block distribution differs, while shallower blocks are generally more HOI-biased and deeper blocks more surrounding-biased.}
  \label{fig:attention-map}
\end{figure}

\section{Preliminary and Motivation}
\label{sec:study}

Before presenting our method, we analyze how temporal memory affects streaming HOI generation under a bounded memory budget. We first study the sink-local trade-off in the standard memory design, and then examine whether different Transformer blocks retrieve historical memory in the same way. These analyses reveal that HOI generation requires not only a proper global sink-local balance, but also block-wise memory organization.

\noindent\textbf{Why does the standard memory organization fail in HOI generation?}
Existing streaming generators usually adopt a fixed sink-local memory, where sink frames preserve early appearance and interaction states, and local frames provide recent motion context. To examine this design in HOI generation, we instantiate it with our base streaming generator distilled on human-object interaction data, whose construction details are provided in Sec.~\ref{sec:implementation-details} and evaluate it on a diagnostic subset of the test set described in Sec.~\ref{sec:datasets}. Under a fixed total memory budget, we vary the sink-local allocation from $(S_{\mathrm{sink}}=1,S_{\mathrm{local}}=11)$ to $(S_{\mathrm{sink}}=11,S_{\mathrm{local}}=1)$ over 19k inferred frames.

As shown in Fig.~\ref{fig:trade-off}(a), small sink memory leads to long-term identity drift and interaction collapse, while small local memory causes action jumps and motion discontinuities. The quantitative results in Fig.~\ref{fig:trade-off}(b,c) confirm this trade-off: increasing sink memory improves Object-CLIP and Subject Consistency, but reducing the local window degrades Interaction Coherence and Motion Smoothness. Here, Interaction Coherence is a VLM-based metric for temporal interaction coherence, with details provided in the supplementary material. These results show that, under bounded memory, the standard sink-local organization cannot jointly maintain long-term HOI stability and short-term motion continuity.

\begin{figure}[tb]
  \centering
  \includegraphics[width=\columnwidth, alt={Placeholder sample figure.}]{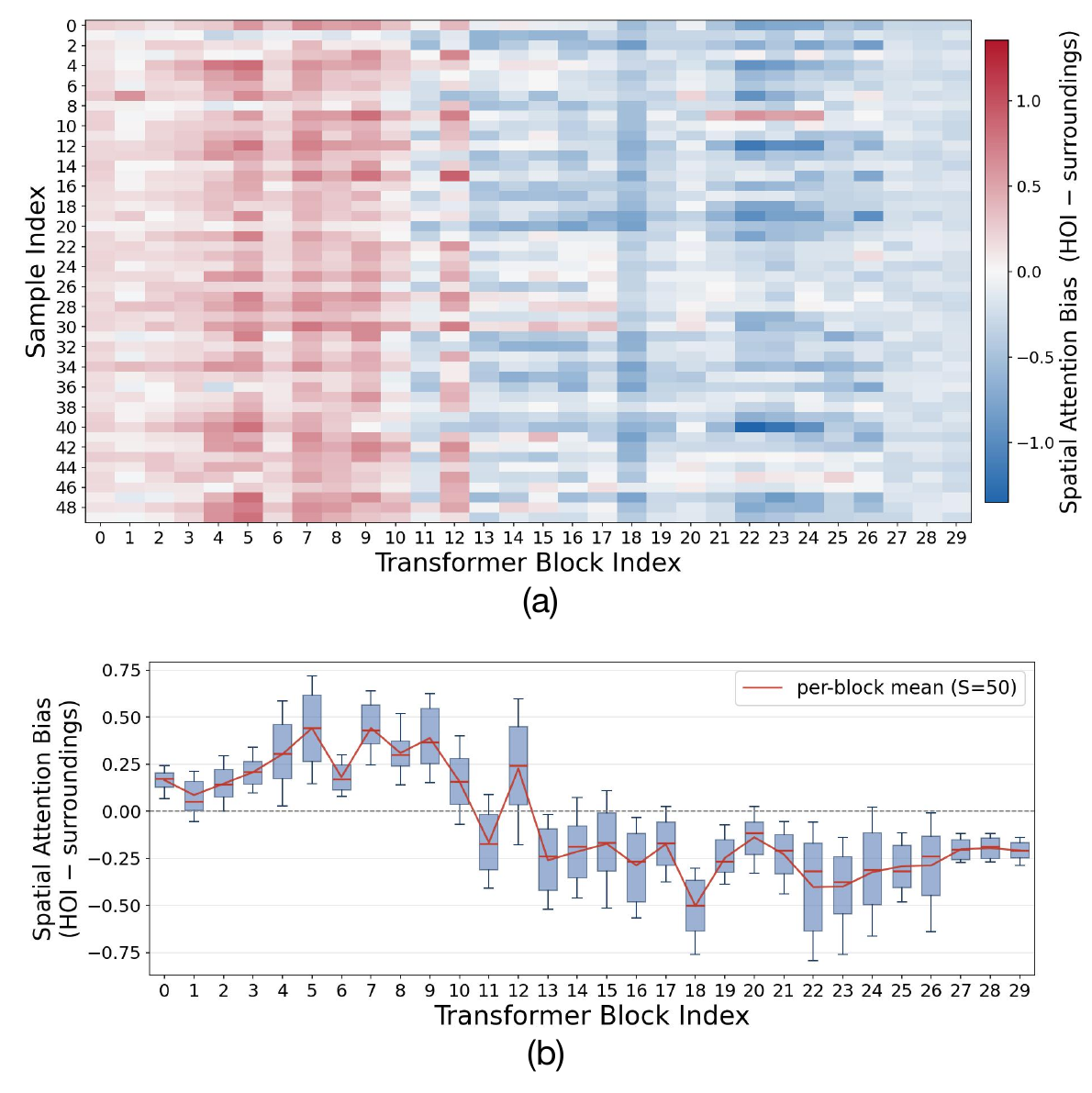}
  \caption{Cross-sample stability of the block-wise spatial attention bias, computed from 50 randomly sampled HOI examples using the same base streaming generator. (a) Sample-wise heatmap; (b) Per-block box plot showing the 25\%–75\% IQR with the median in red and whiskers extending to the 10th and 90th percentiles.}
  \label{fig:sample}
\end{figure}

\noindent\textbf{Do all Transformer blocks need the same memory composition?}
To examine whether the sink-local trade-off can be reduced at the block level, we analyze historical-memory attention in two base models: our base streaming generator and LongLive-5B$^{*}$, our reproduction of LongLive on the Wan2.2-TI2V-5B backbone. We randomly sample 50 HOI examples from HOIGen-1M~\cite{liu2025hoigen} and live e-commerce clips. 
For each block, we visualize the historical-memory attention over the human--object interaction region and the surrounding region. As shown in Fig.~\ref{fig:attention-map}, shallower blocks attend more to the subject, hands, interacted object, and contact area, whereas deeper blocks attend more to the surrounding region. 
We refer to these two tendencies as HOI-biased and surrounding-biased behaviors, respectively.

We further quantify this observation using the profiling procedure later detailed in Sec.~\ref{sec:offline-block-profiling}. Fig.~\ref{fig:sample}(a) shows the sample-block heatmap of spatial attention bias. Most blocks keep a consistent bias direction across samples, forming column-wise red or blue patterns in the heatmap. Fig.~\ref{fig:sample}(b) reports the corresponding per-block distributions; the compact interquartile ranges further suggest that the bias is a stable block-level property rather than a sample-specific artifact. Our subsequent cross-dataset evaluation on the unseen GeoHOI and HOMA test sets (Sec.~\ref{sec:compare-hoi}) further supports the stability of this block-wise memory behavior beyond the sampled observation set.

\noindent\textbf{Motivation for block-wise memory organization.}
The above observation suggests that the sink-local trade-off is not only a global memory-budget problem, but also a block-wise memory-matching problem. In the standard fixed design, all Transformer blocks use the same memory composition, regardless of whether they mainly retrieve historical information for the interaction region or the surrounding region. This uniform design can be suboptimal for HOI generation. For HOI-biased blocks, early sink frames provide important cues for preserving the object identity, subject identity, and hand-object relation. Reducing sink memory in these blocks may weaken long-term interaction states. For surrounding-biased blocks, recent local frames provide more useful cues for short-term visual changes and motion continuity. Allocating too much sink memory to these blocks may reduce the local context needed for smooth temporal propagation. Therefore, the memory composition should be adapted according to the block-wise spatial bias, rather than being shared uniformly across all blocks. This motivates our interaction-aware temporal memory adaptation in Sec.~\ref{sec:method}, where historical memory is organized differently for different Transformer blocks under the same total memory budget.

\section{Method} 
\label{sec:method}

\begin{figure}[tb]
  \centering
  \includegraphics[width=\columnwidth, alt={Placeholder observation figure.}]{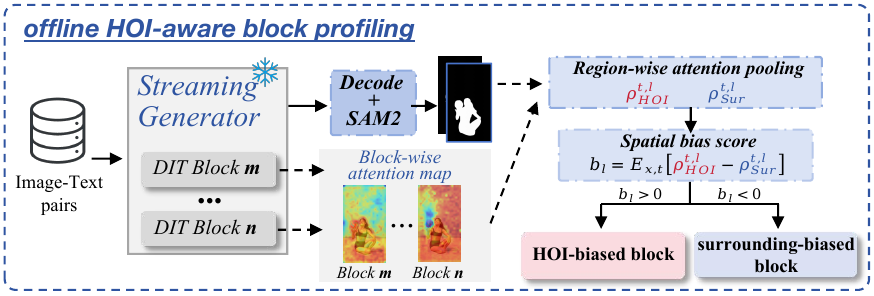}
  \caption{Pipeline of offline HOI-aware block profiling. We utilize SAM2 to pool historical-memory attention across spatial regions, classifying each Transformer block as either HOI-biased or surrounding-biased based on its attention preference.}
  \label{fig:block-profilling}
\end{figure}

\subsection{Overview}
Starting from a few-step streaming generator with a uniform sink-local memory design, StreamHOI adapts historical memory organization at the Transformer-block level for long-duration HOI generation under a fixed memory budget.

\begin{figure*}[tb]
  \centering
  \includegraphics[width=\textwidth, alt={Placeholder observation figure.}]{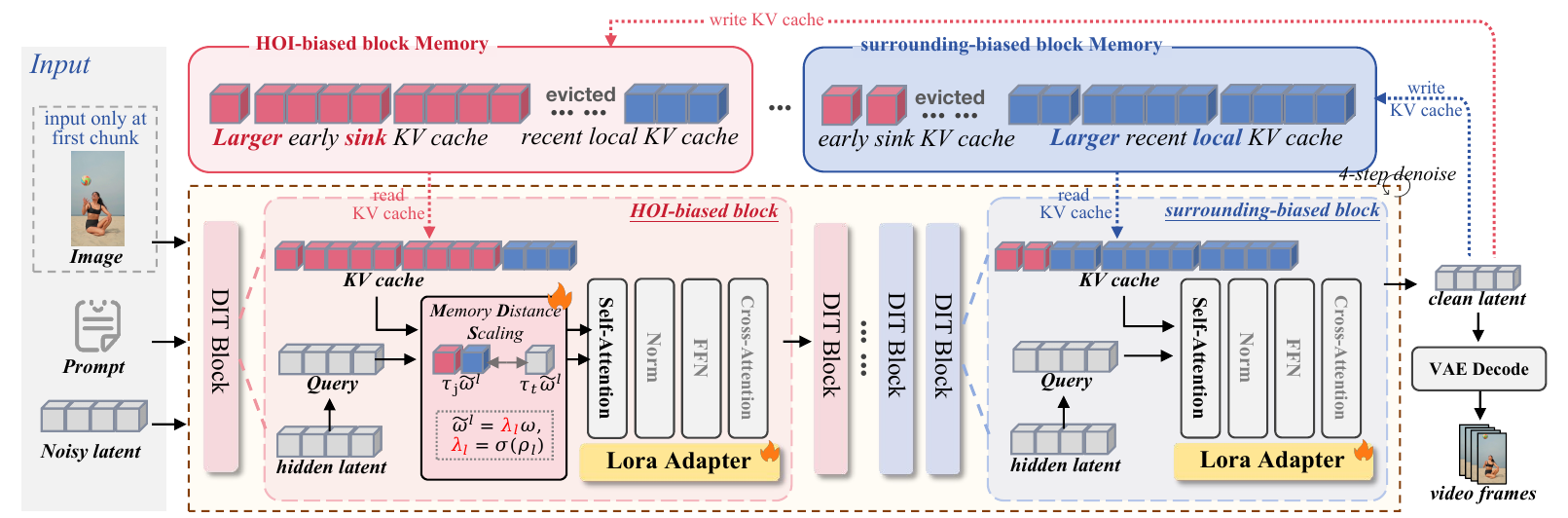}
  \caption{Pipeline of bias-guided memory-specialized training (B-MST). To resolve the sink-local memory trade-off in streaming HOI generation, B-MST specializes memory composition based on the profiled spatial bias: HOI-biased blocks receive more sink memory, coupled with a memory distance scaling (MDS) module, for consistency-oriented use of reliable early interaction states, while surrounding-biased blocks receive more local memory for continuity-oriented use of recent context.}
  \label{fig:pipeline}
\end{figure*}

The pipeline includes offline HOI-aware block profiling and bias-guided memory-specialized training (B-MST). 
Offline HOI-aware block profiling (Fig.~\ref{fig:block-profilling}) identifies the block-wise spatial bias in historical-memory usage. 
To match memory composition with block behavior, B-MST (Fig.~\ref{fig:pipeline}) adapts the generator to a block-specific sink-local memory layout and learns memory distance scaling for long-duration HOI generation. 
The following subsections describe these two components in detail.

\subsection{Offline HOI-aware block profiling}
\label{sec:offline-block-profiling}

Before B-MST, we perform offline HOI-aware block profiling to categorize Transformer blocks according to their historical-memory attention patterns. Some blocks retrieve historical memory more strongly from the HOI regions, including the human, object and contact area, while others rely more on the surrounding region. This provides a direct criterion for block classification: measuring the relative historical-memory attention assigned to the HOI region and the surrounding region.

For offline profiling, we decode latent chunks into RGB frames on a small profiling set and use SAM2~\cite{ravi2024sam2} to obtain the masks of the human subject and the interacted object. The HOI region is defined as the union of the human and object masks, slightly dilated to include local contact areas, and then downsampled to the latent-token resolution. The surrounding region is defined as its complement. These masks are only used during offline profiling and are not required during streaming inference.

For block $l$ at chunk $t$, let $\rho_{\mathrm{hoi}}^{t,l}$ and $\rho_{\mathrm{sur}}^{t,l}$ denote the average historical-memory attention mass assigned to the HOI region and the surrounding region, respectively. The block-wise spatial bias is computed over the profiling videos and chunks:
\begin{equation}
    b_l =
    \frac{1}{|\mathcal{D}_{\mathrm{prof}}||\mathcal{T}_{\mathrm{prof}}|}
    \sum_{x\in\mathcal{D}_{\mathrm{prof}}}
    \sum_{t\in\mathcal{T}_{\mathrm{prof}}}
    \left(
    \rho_{\mathrm{hoi}}^{t,l}(x)
    -
    \rho_{\mathrm{sur}}^{t,l}(x)
    \right),
\end{equation}
where $\mathcal{D}_{\mathrm{prof}}$ denotes the profiling set and $\mathcal{T}_{\mathrm{prof}}$ denotes the profiled chunks. A larger $b_l$ means that block $l$ uses historical memory more strongly for the interaction region, while a smaller $b_l$ indicates stronger reliance on the surrounding region.

The block type is then determined by
\begin{equation}
\mathrm{Type}(l)=
\begin{cases}
\mathrm{HOI\text{-}biased}, & b_l > 0, \\
\mathrm{surrounding\text{-}biased}, & \mathrm{otherwise}.
\end{cases}
\end{equation}
A positive bias means that the block assigns more historical-memory attention to the HOI region than to the surrounding region, and is therefore treated as HOI-biased. Otherwise, the block is treated as surrounding-biased. \textbf{Since the block-wise spatial bias remains stable across interaction scenarios within the same base generator, a small profiling set is sufficient to determine the block types.} 
Implementation details are provided in the supplementary material.

\subsection{Bias-guided memory-specialized training}
\label{sec:B-MST}

After obtaining the HOI-aware block types through offline profiling, we train the streaming generator with a block-specialized memory layout. The goal is to align the temporal memory composition of each Transformer block with its spatial memory preference. HOI-biased blocks are encouraged to use more reliable early states for preserving interaction consistency, while surrounding-biased blocks are encouraged to use richer recent context for maintaining local continuity. We refer to this second-stage training strategy as \emph{bias-guided memory-specialized training} (B-MST).

Let $B$ denote the total memory budget for each Transformer block. In the uniform streaming generator, every block uses the same sink-local allocation $(s_0,w_0)$, where $B=s_0+w_0$. B-MST instead assigns a block-specific allocation $(s_l,w_l)$ according to the spatial bias score $b_l$ obtained in Sec.~\ref{sec:offline-block-profiling}. We first normalize the block-wise bias as
\begin{equation}
    \hat{b}_l =
    \frac{b_l}{\max_k |b_k|+\epsilon},
\end{equation}
where $\epsilon$ is a small constant for numerical stability. The sink size of block $l$ is then defined as
\begin{equation}
    s_l =
    \mathrm{Clip}
    \left(
    \mathrm{Round}
    \left(
    s_0 + B \hat{b}_l
    \right),
    s_{\min},
    B-w_{\min}
    \right),
\end{equation}
and the local-window size is
\begin{equation}
    w_l = B - s_l .
\end{equation}
Here, $s_{\min}$ and $w_{\min}$ ensure that each block retains a minimum number of sink and local-window frames. This allocation preserves the same total memory budget for every block. A positive $\hat{b}_l$ increases the sink size for HOI-biased blocks, allowing them to retain more early latent frames that contain stable object appearance, contact relations, and manipulation states. A negative $\hat{b}_l$ increases the local-window size for surrounding-biased blocks, allowing them to condition on more recent latent frames for local motion and scene continuity.

At chunk $t$, the block-specific KV memory is constructed as
\begin{equation}
\left\{
\begin{aligned}
\mathcal{M}_{l}^{t}
&=
\mathrm{Concat}
\left(
\mathcal{M}_{l,\mathrm{sink}}^{t},
\mathcal{M}_{l,\mathrm{local}}^{t}
\right), \\
\mathcal{M}_{l,\mathrm{sink}}^{t}
&=
\mathrm{KV}_{l}
\left(
\{z_0,\ldots,z_{s_l-1}\}
\right), \\
\mathcal{M}_{l,\mathrm{local}}^{t}
&=
\mathrm{KV}_{l}
\left(
\{z_{tm-w_l},\ldots,z_{tm-1}\}
\right).
\end{aligned}
\right.
\end{equation}
Thus, B-MST changes only the memory composition of each block, without increasing the total KV-cache budget.

\noindent\textbf{Memory distance scaling.}
Although sink memory preserves early interaction states, its key tokens are temporally far from the current chunk during long streaming generation. Under a standard temporal RoPE, the relative rotation between a query at temporal position $\tau_t$ and a key at $\tau_j$ is proportional to $(\tau_t-\tau_j)\boldsymbol{\omega}$, where $\boldsymbol{\omega}$ is the temporal RoPE frequency. Because $\tau_t-\tau_j$ is large for sink tokens, the resulting rapid phase variation can weaken their query--key correlation and reduce the contribution of early interaction states. To alleviate this issue, B-MST introduces a learnable memory distance scaling mechanism for HOI-biased blocks that acts on the temporal RoPE frequency.

For a HOI-biased block $l$, we rescale its temporal RoPE frequency by a learnable block-specific factor:
\begin{equation}
    \tilde{\boldsymbol{\omega}}^{l}
    =
    \lambda_l\boldsymbol{\omega},
    \qquad
    \lambda_l = \sigma(\rho_l),
\end{equation}
where $\rho_l\in\mathbb{R}$ is a learnable scalar and $\sigma(\cdot)$ is the sigmoid function. For surrounding-biased blocks, we fix $\lambda_l\equiv 1$ so that the original temporal distance is preserved and no additional parameter is introduced.

The phase perturbation induced by this scaling is
\begin{equation}
    \Delta\boldsymbol{\phi}^{l}_{tj}
    =
    (1-\lambda_l)(\tau_t-\tau_j)\boldsymbol{\omega}.
\end{equation}
This expression shows that MDS is not a hard sink-only re-indexing operation: since the temporal RoPE frequency is rescaled at the block level, all query--key temporal distances in the corresponding HOI-biased block are affected. However, its effect is inherently distance-dependent. Local-window keys have bounded temporal distances to the current chunk, whereas sink keys become increasingly distant as streaming generation proceeds. Therefore, the same scaling factor produces a much stronger phase correction for distant sink tokens than for recent local tokens.

MDS is intentionally applied only to HOI-biased blocks. These blocks are assigned larger sink memories by B-MST and require stronger access to early interaction anchors, such as contact relations, relative poses, and object states. Surrounding-biased blocks keep the original RoPE frequency, preserving their recent-context behavior for motion and scene continuity.

\noindent\textbf{Training objective.}
B-MST is conducted as a lightweight second-stage training procedure. We freeze the pretrained streaming generator and optimize only LoRA adapters $\phi$ inserted into the Transformer blocks, together with the distance-scaling parameters $\{\rho_l\}$ for HOI-biased blocks. Given the reference human image $I_H$, text prompt $P$, latent noise $\epsilon^t$, and block-specific memories $\{\mathcal{M}_l^t\}_{l=0}^{L-1}$, the generator produces the current latent chunk:
\begin{equation}
    \hat{Z}^t =
    G_{\theta,\phi}
    \left(
    I_H, P, \epsilon^t, \{\mathcal{M}_l^t\}_{l=0}^{L-1}
    \right),
\end{equation}
where $\theta$ denotes the frozen generator parameters.

We use the DMD distillation objective:
\begin{equation}
    \mathcal{L}_{\mathrm{B-MST}}
    =
    \mathcal{L}_{\mathrm{DMD}}
    +
    \lambda_{\mathrm{c}}
    \mathcal{L}_{\mathrm{critic}}.
\end{equation}
Here, $\mathcal{L}_{\mathrm{DMD}}$ updates the LoRA-augmented generator by matching the generated distribution to the real-score model, while $\mathcal{L}_{\mathrm{critic}}$ updates the fake-score critic with a denoising objective on generated samples. 

Different from the standard DMD training that rolls out the generator with a uniform sink-local memory for all Transformer blocks, B-MST performs generator rollouts and score evaluations with the bias-guided memory layout, so the distillation gradients are computed under block-specialized historical contexts.

Through B-MST, the streaming generator learns to use historical memory according to the functional preference of each block. HOI-biased blocks are strengthened toward consistency-oriented memory use, while surrounding-biased blocks are strengthened toward continuity-oriented memory use. During inference, we use the profiled block-specific memory layout and the learned distance-scaling coefficients without introducing additional model branches or increasing the total memory budget.

\section{Experiments}
\subsection{Experimental Settings}
\subsubsection{Implementation Details}
\label{sec:implementation-details}
We use Wan2.2-TI2V-5B as the base video diffusion model and generate videos at a resolution of $704 \times 1280$. 
We first construct a chunk-wise few-step streaming generator through teacher-forcing causal training, causal ODE distillation, and DMD distillation, and then apply the proposed bias-guided memory-specialized training (B-MST) for interaction-aware temporal memory adaptation. 
All stages are implemented in a chunk-wise streaming manner, where each chunk contains 4 latent frames. 
In the teacher-forcing causal training stage, we train an autoregressive diffusion model for 13K steps on 100K video clips. 
We then use the trained autoregressive model as the teacher to sample 10K causal ODE trajectories, and perform causal ODE distillation for 5K steps to obtain a 4-step generator. 
In the DMD distillation stage, we adopt a uniform sink-local memory configuration with $s=4$ sink latent frames and $w=8$ local-window latent frames, and train the streaming generator for 300 steps. 
Finally, we apply B-MST for 700 steps under the profiled block-specific memory layout. 
The LoRA rank and alpha are both set to 256. 
All training experiments are conducted on 32 80GB GPUs.

\subsubsection{Dataset}
\label{sec:datasets}

\noindent\textbf{Training dataset.} We train StreamHOI on a 100K video dataset. The dataset consists of 80K internal live e-commerce video clips and 20K HOI clips sampled from HOIGen-1M\cite{liu2025hoigen}. The internal videos contain diverse human demonstrations and object manipulation scenarios in live e-commerce settings, while the HOIGen-1M subset provides additional human--object interaction examples with broader interaction categories.

\begin{table*}[tb]
\centering
\caption{Quantitative comparison with representative methods under different temporal settings on the main test set. LC-Avatar denotes Longcat-Video-Avatar 1.5. Our 5B model substantially improves HOI-specific metrics while maintaining high general visual quality, ranking second only to the 13.6B LC-Avatar on most general metrics.}
\label{tab:main_comparison}
\resizebox{\textwidth}{!}{
\begin{tabular}{llcccccccc}
\toprule
\multirow{2}{*}{} & \multirow{2}{*}{\textbf{Method}}
& \multicolumn{4}{c}{\textbf{HOI-specific Metrics}}
& \multicolumn{4}{c}{\textbf{General Metrics}} \\
\cmidrule(lr){3-6} \cmidrule(lr){7-10}
& & Obj-CLIP$\uparrow$
& InternVL$_O$$\uparrow$
& InternVL$_H$$\uparrow$
& InternVL$_I$$\uparrow$
& \makecell{Subject\\Cons.$\uparrow$}
& \makecell{Background\\Cons.$\uparrow$}
& \makecell{Motion\\Smooth.$\uparrow$}
& \makecell{Aesthetic\\Quality$\uparrow$} \\
\midrule

\multirow{8}{*}{\rotatebox[origin=c]{90}{5-second}}
& Causal Forcing-1.3B     & 0.7104 & 0.823 & 0.921 & 0.836 & 0.9688 & 0.9665 & 0.9821 & 0.5298 \\
& LongLive-1.3B           & 0.7261 & 0.821 & 0.908 & 0.842 & 0.9683 & 0.9672 & 0.9815 & 0.5281 \\
& Causal Forcing-5B$^{*}$ & 0.8032 & 0.869 & 0.937 & 0.860 & 0.9699 & 0.9724 & 0.9833 & 0.5493 \\
& LongLive-5B$^{*}$       & 0.7927 & 0.862 & \underline{0.994} & 0.857 & 0.9742 & 0.9778 & 0.9869 & 0.5490 \\
& Deep Forcing-5B         & 0.8181 & 0.885 & 0.982 & 0.864 & 0.9729 & 0.9780 & 0.9877 & \underline{0.5503} \\
& LC-Avatar-13.6B         & \underline{0.8539} & \underline{0.899} & \textbf{0.995} & \underline{0.889} & \textbf{0.9879} & \textbf{0.9893} & \textbf{0.9969} & 0.5482 \\
& InteractAvatar-10B      & 0.7688 & 0.851 & 0.982 & 0.884 & 0.9760 & 0.9789 & 0.9893 & 0.5426 \\
\rowcolor{orange!12}
& Ours-5B                 & \textbf{0.8621} & \textbf{0.909} & 0.993 & \textbf{0.901} & \underline{0.9782} & \underline{0.9801} & \underline{0.9934} & \textbf{0.5558} \\
\midrule

\multirow{8}{*}{\rotatebox[origin=c]{90}{30-second}}
& Causal Forcing-1.3B     & 0.6033 & 0.681 & 0.794 & 0.704 & 0.9486 & 0.9599 & 0.9812 & 0.5139 \\
& LongLive-1.3B           & 0.6109 & 0.631 & 0.802 & 0.715 & 0.9495 & 0.9622 & 0.9808 & 0.5118 \\
& Causal Forcing-5B$^{*}$ & 0.6607 & 0.782 & 0.831 & 0.782 & 0.9586 & 0.9681 & 0.9827 & 0.5211 \\
& LongLive-5B$^{*}$       & 0.6586 & 0.789 & 0.824 & 0.790 & 0.9701 & 0.9766 & 0.9845 & 0.5234 \\
& Deep Forcing-5B         & 0.6936 & 0.805 & 0.816 & 0.801 & 0.9692 & 0.9773 & 0.9858 & 0.5269 \\
& LC-Avatar-13.6B         & \underline{0.8244} & \underline{0.863} & \underline{0.951} & \underline{0.866} & \textbf{0.9869} & \textbf{0.9891} & \textbf{0.9966} & \textbf{0.5438} \\
& InteractAvatar-10B      & 0.7492 & 0.827 & 0.924 & 0.822 & 0.9709 & 0.9785 & 0.9872 & 0.5377 \\
\rowcolor{orange!12}
& Ours-5B                 & \textbf{0.8496} & \textbf{0.882} & \textbf{0.958} & \textbf{0.874} & \underline{0.9751} & \underline{0.9802} & \underline{0.9923} & \underline{0.5431} \\
\midrule

\multirow{8}{*}{\rotatebox[origin=c]{90}{60-second}}
& Causal Forcing-1.3B     & 0.4930 & 0.537 & 0.688 & 0.613 & 0.9480 & 0.9582 & 0.9796 & 0.4998 \\
& LongLive-1.3B           & 0.5164 & 0.548 & 0.707 & 0.626 & 0.9488 & 0.9603 & 0.9821 & 0.4981 \\
& Causal Forcing-5B$^{*}$ & 0.5882 & 0.687 & 0.792 & 0.702 & 0.9577 & 0.9669 & 0.9828 & 0.5203 \\
& LongLive-5B$^{*}$       & 0.5937 & 0.688 & 0.808 & 0.712 & 0.9640 & 0.9760 & 0.9842 & 0.5193 \\
& Deep Forcing-5B         & 0.6334 & 0.703 & 0.831 & 0.745 & 0.9659 & 0.9731 & 0.9857 & 0.5251 \\
& LC-Avatar-13.6B         & \underline{0.8156} & \underline{0.842} & \underline{0.940} & \underline{0.834} & \textbf{0.9846} & \textbf{0.9880} & \textbf{0.9963} & \textbf{0.5431} \\
& InteractAvatar-10B      & 0.7344 & 0.808 & 0.923 & 0.791 & 0.9693 & 0.9781 & 0.9878 & 0.5372 \\
\rowcolor{orange!12}
& Ours-5B                 & \textbf{0.8298} & \textbf{0.857} & \textbf{0.942} & \textbf{0.868} & \underline{0.9734} & \underline{0.9796} & \underline{0.9902} & \underline{0.5429} \\
\bottomrule
\end{tabular}
}
\end{table*}
\begin{table*}[tb]
\centering
\caption{Comparison with bidirectional HOI methods under the 5-second setting. Uni-DiT denotes UniAnimate-DiT. Our method achieves competitive performance without extra object-information injection.}
\label{tab:compare-5s}
\resizebox{\textwidth}{!}{%
\begin{tabular}{lcccccccc}
\toprule
\multirow{2}{*}{\textbf{Method}}
& \multicolumn{4}{c}{\textbf{HOI-specific Metrics}}
& \multicolumn{4}{c}{\textbf{General Metrics}} \\
\cmidrule(lr){2-5}
\cmidrule(lr){6-9}
& Obj-CLIP$\uparrow$
& InternVL$_O$$\uparrow$
& InternVL$_H$$\uparrow$
& InternVL$_I$$\uparrow$
& \makecell{Subject\\Cons.$\uparrow$}
& \makecell{Background\\Cons.$\uparrow$}
& \makecell{Motion\\Smooth.$\uparrow$}
& \makecell{Aesthetic\\Quality$\uparrow$} \\
\midrule

\rowcolor{gray!15}
\multicolumn{9}{c}{\textbf{GeoHOI test set}} \\
Uni-DiT-14B        & 0.8786 & 0.818 & 0.945 & 0.848 & 0.9814 & 0.9858 & 0.9948 & 0.5274 \\
VACE-14B           & 0.8516 & 0.736 & 0.962 & 0.825 & \textbf{0.9831} & \underline{0.9863} & \underline{0.9951} & 0.5391 \\
GeoHOI-5B          & \textbf{0.9015} & \underline{0.882} & \underline{0.963} & \textbf{0.898} & 0.9810 & 0.9860 & 0.9923 & \underline{0.5396} \\
\rowcolor{orange!12}
\textbf{Ours-5B}   & \underline{0.8993} & \textbf{0.914} & \textbf{0.980} & \underline{0.892} & \underline{0.9824} & \textbf{0.9879} & \textbf{0.9957} & \textbf{0.5483} \\

\midrule

\rowcolor{gray!15}
\multicolumn{9}{c}{\textbf{HOMA test set}} \\
HUMO-17B           & 0.8560 & 0.758 & 0.926 & 0.769 & \underline{0.9792} & \textbf{0.9891} & \textbf{0.9940} & 0.5100 \\
VACE-14B           & 0.8572 & 0.880 & \textbf{0.997} & 0.842 & 0.9788 & \underline{0.9878} & \underline{0.9900} & 0.5350 \\
HOMA-13B           & \textbf{0.8803} & \underline{0.887} & 0.983 & \underline{0.885} & 0.9774 & 0.9860 & 0.9810 & \textbf{0.5860} \\
\rowcolor{orange!12}
\textbf{Ours-5B}   & \underline{0.8790} & \textbf{0.905} & \underline{0.984} & \textbf{0.896} & \textbf{0.9806} & 0.9869 & 0.9830 & \underline{0.5380} \\

\bottomrule
\end{tabular}%
}
\end{table*}

\noindent\textbf{Evaluation dataset.}
Existing HOI test sets mainly target short offline video generation and cover limited interaction scenarios. 
We therefore construct a main test set under diverse real-world interaction scenarios, containing 400 HOI cases and resulting in 384k inferred frames. Specifically, it includes 100 live e-commerce videos and 300 videos selected from HOIGen-1M, all excluded from training. 
To characterize its scenario diversity, we group the videos into coarse interaction categories according to their captions, as summarized in Table~\ref{tab:hoigen-testset}.

In addition, we use GeoHOI-testset~\cite{xu2026geohoi} and HOMA-testset~\cite{huang2025hyvideohoma} for comparison with existing bidirectional HOI generation methods. 
These two test sets are tied to their corresponding methods and provide standardized evaluation prompts or conditions for short-video HOI generation. Each test set contains 100 examples. 
Therefore, we use them only for 5-second comparison with bidirectional HOI baselines.

\begin{figure*}[tb]
  \centering
  \includegraphics[width=\textwidth, alt={Placeholder observation figure.}]{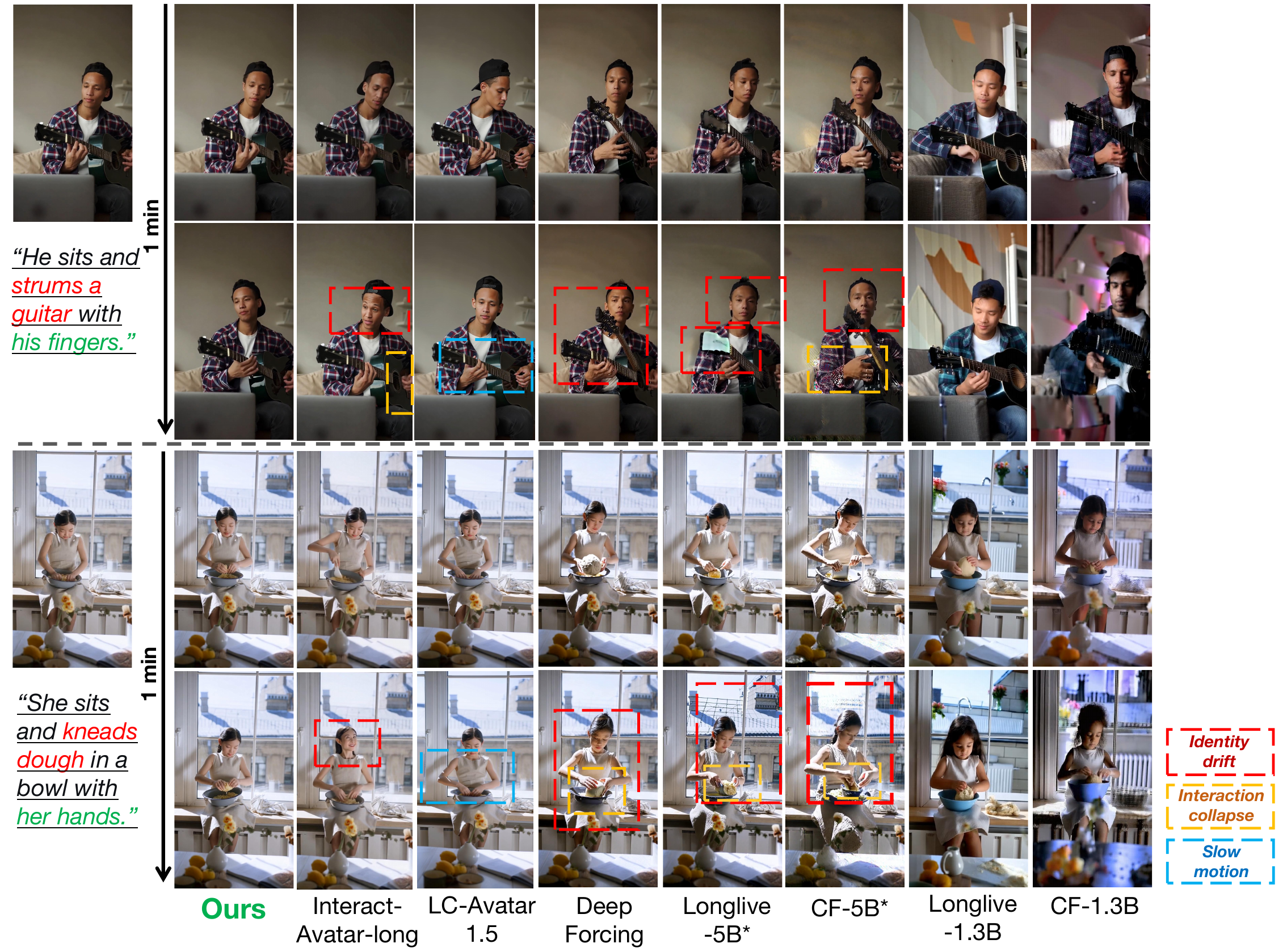}
  \caption{Qualitative comparison against state-of-the-art long-video generation frameworks. While baseline methods exhibit severe identity drift, object distortion, or temporal inconsistencies over extended contexts, our approach preserves robust subject-object fidelity and coherent interaction dynamics. Please refer to the supplementary materials for the full video sequences.}
  \label{fig:main-comparison}
\end{figure*}

\begin{figure}[tb]
  \centering
  \includegraphics[width=\columnwidth, alt={Placeholder longcat comparison.}]{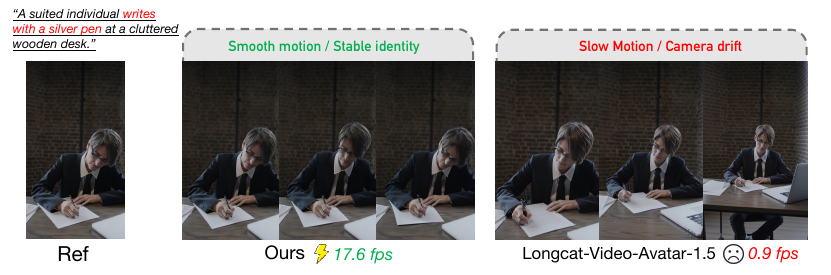}
  \caption{Further comparison with Longcat-Video-Avatar 1.5 under the 60-second setting. Video results can be found in the supplementary materials.}
  \label{fig:longcat comparison}
\end{figure}

\begin{table}[tb]
\centering
\caption{Interaction category distribution of the main evaluation test set.}
\label{tab:hoigen-testset}
\begin{tabular}{lc}
\toprule
\textbf{Interaction Category} & \textbf{\#Videos} \\
\midrule
Digital device interaction & 46 \\
Food preparation & 44 \\
Art / craft creation & 40 \\
Footwear handling & 40 \\
Object demonstration & 40 \\
Wearable / grooming adjustment & 34 \\
Sports / mobility activity & 31 \\
Gesture / explanation & 30 \\
Household / plant / pet care & 29 \\
Study / tool use & 24 \\
Music / dance performance & 23 \\
Food / drink serving & 19 \\
\midrule
Total & 400 \\
\bottomrule
\end{tabular}
\end{table}

\subsubsection{Comparison Methods}
\label{sec:comparison-methods}

We compare StreamHOI with two groups of baselines. 
For long-video generation methods, we evaluate all methods under the 5-, 30-, and 60-second settings to measure both short-term quality and long-duration stability. 
For bidirectional HOI generation methods, we evaluate them under the 5-second setting, since these methods are mainly designed for short offline video generation.

\noindent\textbf{Long-video generation methods.}
First, we compare with representative streaming video generation methods, including Causal Forcing (frame-wise)~\cite{zhu2026causal}, LongLive~\cite{yang2025longlive}, and Deep Forcing~\cite{yi2025deep}. 
Since Causal Forcing and LongLive only release 1.3B model weights, we reproduce them on the Wan2.2-TI2V-5B backbone using our training data and closely matched training configurations, denoted as Causal Forcing-5B$^{*}$ and LongLive-5B$^{*}$. 
Deep Forcing~\cite{yi2025deep} is a training-free long-video generation method; we apply it to our baseline model without B-MST for comparison. 
We also compare with InteractAvatar-long~\cite{zhang2026making}, an offline long-video HOI generation method based on sequential clip generation. 
In addition, we include LongCat-Video-Avatar 1.5~\cite{team2026longcat}, a 13.6B few-step avatar video generation method. 
Since LongCat-Video-Avatar 1.5 additionally takes audio as input, we use a silent audio track to keep the comparison under the same image-and-text-driven setting. 
All these methods are evaluated on the main test set.

\noindent\textbf{Bidirectional HOI methods.} Second, we compare with bidirectional HOI video generation methods on HOI test sets under the 5-second setting, as these baselines are mainly designed for short video generation. 
Since StreamHOI only takes a reference image and a text prompt as input, we convert the hand-pose and object-trajectory conditions provided by the datasets into textual descriptions using Qwen2.5-VL~\cite{Qwen2.5-VL}, and append them to the prompts for our method. Therefore, this experiment focuses on comparing the final HOI generation quality on the same test cases, rather than strict condition-following accuracy.
On GeoHOI-testset, we compare with UniAnimate-DiT\cite{wang2025unianimatedit}, VACE\cite{jiang2025vace}, and GeoHOI\cite{xu2026geohoi}. 
UniAnimate-DiT generates human videos from a person image and a pose sequence based on a Wan-14B backbone. 
VACE is a general video editing framework built on a 14B video model and supports text-controlled generation and editing. 
GeoHOI improves HOI video generation by enhancing object awareness and geometric interaction control. 
On HOMA-testset, we compare with HUMO\cite{chen2025humo}, VACE, and HOMA\cite{huang2025hyvideohoma}. 
HUMO supports video generation conditioned on multiple reference images. 
HOMA introduces weak motion guidance for HOI human video generation.

\subsubsection{Metrics}
We evaluate our method using several established metrics. Object-CLIP\cite{xu2024anchorcrafter} is used to evaluate
the object consistency by computing the CLIP\cite{radford2021learning} cosine similarity of the objects between the input image and the generated videos within the segmentation. 
Since traditional metrics struggle to capture fine-grained physical contact and interaction evolution, we further use InternVL3-38B~\cite{zhu2025internvl3} as a structured evaluator. It performs a multi-dimensional binary assessment on sampled frames to evaluate object fidelity, human quality, and interaction plausibility.
Details of the metrics are provided in the supplementary materials. 
Subject Consistency, Background Consistency, Motion smoothness and Aesthetic Quality extracted from Vbench\cite{huang2024vbench} are employed to measure general video quality. 
We also report FPS and Latency as efficiency metrics in Sec.~\ref{sec:runtime}.

\subsection{Comparison Results}
\subsubsection{Comparison with long-video generation methods}

According to Table~\ref{tab:main_comparison}, our method consistently achieves the best results across nearly all quality metrics under different temporal settings. For HOI-specific evaluation, our method obtains the highest Obj-CLIP scores at 5, 30, and 60 seconds, showing better object appearance preservation than representative baselines. The InternVL-based evaluation further shows that our method better maintains human and object states while producing more reliable hand-object interactions. For general video quality, our method also achieves strong Subject Consistency and Background Consistency, indicating stable foreground subjects and scenes over time. Meanwhile, the competitive Motion Smoothness and Aesthetic Quality scores show that our method maintains smooth motion and high visual quality across different video lengths. Detailed per-category results and additional Qwen3-VL-based evaluations~\cite{Qwen3-VL} are provided in the supplementary material; \textbf{the Qwen3-VL-based results assess cross-evaluator robustness and show broadly consistent performance trends with InternVL3-38B.}

A key advantage of our method is its strong long-duration generation ability. As the generation length increases from 5 seconds to 60 seconds, most streaming baselines show clear performance drops, especially on Obj-CLIP and InternVL scores. This suggests that they gradually lose early object and interaction states during long autoregressive generation. In contrast, our method keeps high scores even at 60 seconds, with only a small decrease. The qualitative results in Fig.~\ref{fig:main-comparison} show a similar trend. Streaming baselines often suffer from object drift, weak hand-object contact, inconsistent object appearance, and degraded motion after long generation. Our method better maintains human identity, object appearance, background stability, and physically plausible interaction dynamics throughout the video.

We further show qualitative comparisons with LongCat-Video-Avatar 1.5~\cite{team2026longcat}, a 13.6B few-step avatar video generation method. As shown in Fig.~\ref{fig:longcat comparison}, LongCat-Video-Avatar 1.5 tends to produce slow motion and shows camera drift over long generation. In contrast, our method maintains more natural motion dynamics, more stable human identity throughout long videos.

\begin{figure}[tb]
  \centering
  \includegraphics[width=\columnwidth, alt={Placeholder geohoi-testset comparison.}]{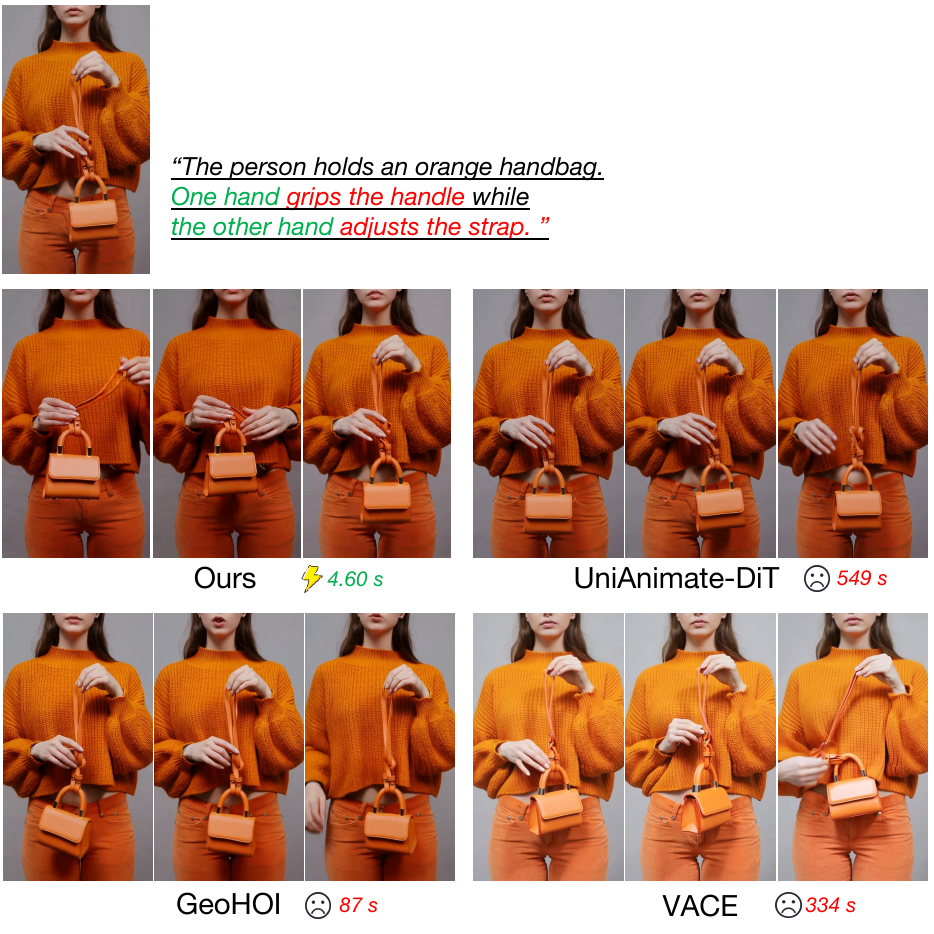}
  \caption{Comparison with SOTAs on the GeoHOI-testset under the 5-second setting. Video results can be found in the supplementary materials.}
  \label{fig:geohoi-testset-comparison}
\end{figure}

\begin{figure}[tb]
  \centering
  \includegraphics[width=\columnwidth, alt={Placeholder homa-testset comparison.}]{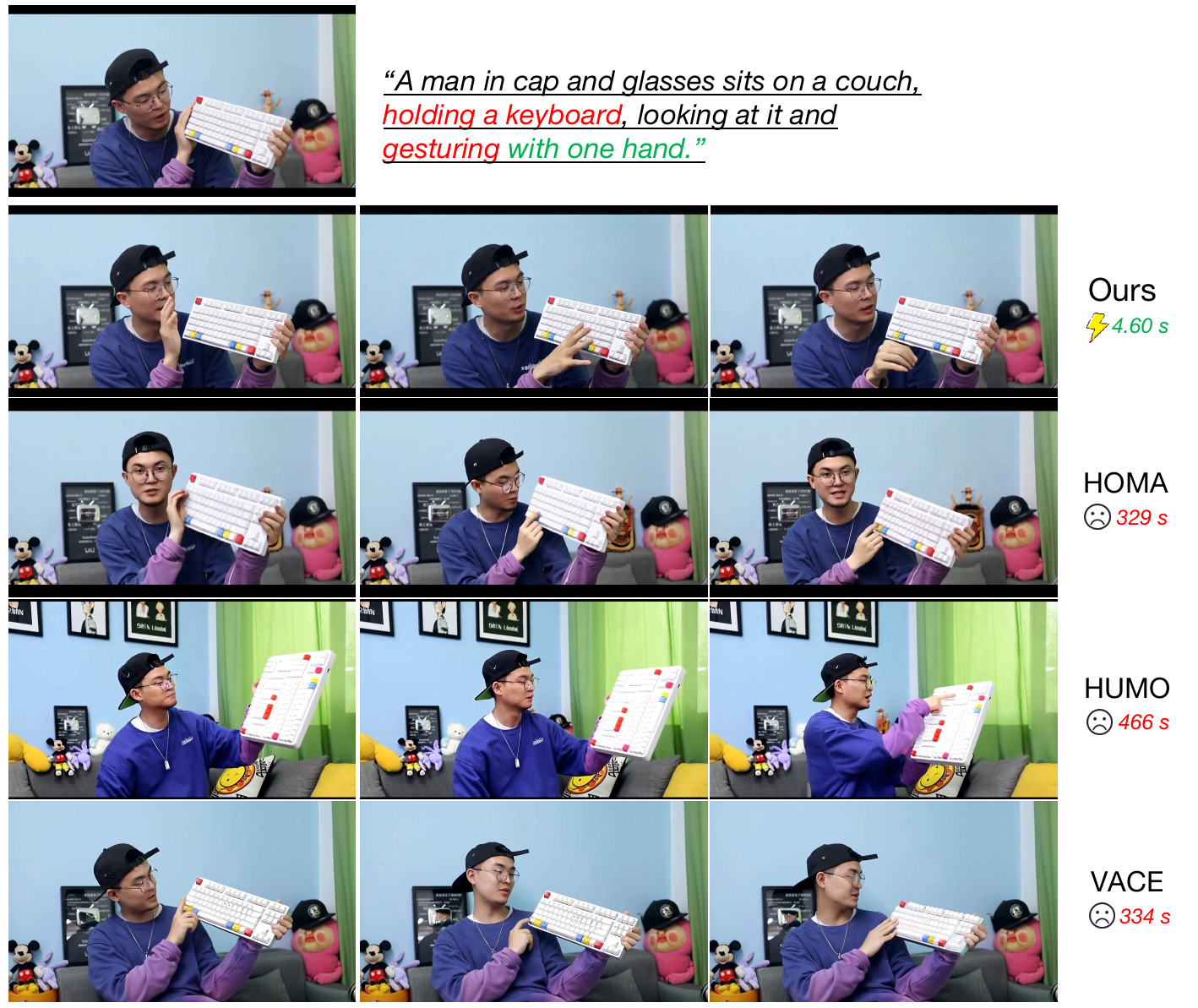}
  \caption{Comparison with SOTAs on the HOMA-testset under the 5-second setting. Video results can be found in the supplementary materials.}
  \label{fig:homa-testset-comparison}
\end{figure}

\subsubsection{Comparison with bidirectional HOI methods}
\label{sec:compare-hoi}
As shown in Table~\ref{tab:compare-5s}, our model achieves highly competitive results compared with recent bidirectional HOI generation methods. On the HOI-specific metrics, our method obtains higher Obj-CLIP than UniAnimate-DiT, VACE, and HUMO, showing stronger object appearance preservation. Without any additional object condition injection, our method still achieves comparable Obj-CLIP performance with GeoHOI and HOMA. More importantly, our method obtains very competitive InternVL$_O$ and InternVL$_I$ scores, indicating that it can effectively maintain object consistency and produce reasonable human--object interactions. The InternVL$_H$ result also shows that our method preserves the human appearance well during generation. In addition, our method achieves strong performance on general video metrics, demonstrating that the proposed streaming framework can jointly support high-quality HOI generation and strong overall video quality. 
\textit{Crucially, achieving such competitive performance on the completely unseen GeoHOI and HOMA test sets firmly validates the cross-dataset generalization of our offline block profiling design.}

As shown in Fig.~\ref{fig:geohoi-testset-comparison}, we present qualitative comparisons on the GeoHOI-testset. Our method preserves the object appearance well and produces a reasonable interaction process. Even without pose guidance or object-driven control, our method can naturally adjust the strap according to the interaction context. In contrast, UniAnimate-DiT and GeoHOI require additional pose injection, yet their generated results appear less natural, while VACE shows clear color-tone changes in the generated video. As shown in Fig.~\ref{fig:homa-testset-comparison}, we further present qualitative results on the HOMA-testset. Our method preserves the face identity well and generates natural contact between the hands and the keyboard. Other methods show more visible artifacts: the generated faces are often distorted, HOMA fails to follow the instruction that the person should look at the keyboard, HUMO changes the keyboard appearance, and VACE changes the shape of the hat. These qualitative results further show that our method can maintain human identity, object appearance, and interaction plausibility. It is worth noting that our method generates a 5-second video in only 4.6 seconds, which is much faster than existing bidirectional HOI methods.

\begin{table*}[tb]
\centering
\caption{Ablation study under the 60-second setting. We ablate bias-guided memory-specialized training (B-MST), memory distance scaling (MDS), and the block grouping strategy. Uniform-Mem FT denotes uniform-memory fine-tuning, an equal-step LoRA fine-tuning baseline that uses the same 700 LoRA fine-tuning steps and the same total KV-cache budget as B-MST, while keeping the original uniform sink-local memory layout.}
\label{tab:main_ablation}
\resizebox{\textwidth}{!}{
\begin{tabular}{lcccccccc}
\toprule
\multirow{2}{*}{\textbf{Method}}
& \multicolumn{4}{c}{\textbf{HOI-specific Metrics}}
& \multicolumn{4}{c}{\textbf{General Metrics}} \\
\cmidrule(lr){2-5} \cmidrule(lr){6-9}
& Obj-CLIP$\uparrow$
& InternVL$_O$$\uparrow$
& InternVL$_H$$\uparrow$
& InternVL$_I$$\uparrow$
& \makecell{Subject\\Cons.$\uparrow$}
& \makecell{Background\\Cons.$\uparrow$}
& \makecell{Motion\\Smooth.$\uparrow$}
& \makecell{Aesthetic\\Quality$\uparrow$} \\
\midrule
Ours
& \textbf{0.8298} & \textbf{0.857} & \textbf{0.942} & \textbf{0.868} & \textbf{0.9734} & \textbf{0.9796} & \textbf{0.9902} & \textbf{0.5429} \\
w/ fixed MDS
& 0.8204 & 0.840 & 0.923 & 0.864 & 0.9730 & 0.9792 & 0.9885 & 0.5366 \\
w/o MDS
& 0.7519 & 0.781 & 0.880 & 0.774 & 0.9647 & 0.9759 & 0.9883 & 0.5279 \\
w/o B-MST
& 0.6028 & 0.682 & 0.836 & 0.701 & 0.9583 & 0.9748 & 0.9851 & 0.5236 \\
Uniform-Mem FT 
& 0.6093 & 0.675 & 0.859 & 0.688 & 0.9592 & 0.9744 & 0.9855 & 0.5269 \\
\midrule
Reverse Grouping
& 0.5744 & 0.671 & 0.808 & 0.662 & 0.9579 & 0.9747 & 0.9839 & 0.5231 \\
Random Grouping
& 0.5939 & 0.686 & 0.821 & 0.724 & 0.9601 & 0.9732 & 0.9852 & 0.5282 \\
\bottomrule
\end{tabular}
}
\end{table*}
\begin{table*}[tb]
\centering
\caption{Generalization of Bias-Guided Memory-Specialized Training.}
\label{tab:add_cache_allocation_to_longlive}
\resizebox{\textwidth}{!}{
\begin{tabular}{lcccccccc}
\toprule
\multirow{2}{*}{\textbf{Method}}
& \multicolumn{4}{c}{\textbf{HOI-specific Metrics}}
& \multicolumn{4}{c}{\textbf{General Metrics}} \\
\cmidrule(lr){2-5} \cmidrule(lr){6-9}
& Obj-CLIP$\uparrow$
& InternVL$_O$$\uparrow$
& InternVL$_H$$\uparrow$
& InternVL$_I$$\uparrow$
& \makecell{Subject\\Cons.$\uparrow$}
& \makecell{Background\\Cons.$\uparrow$}
& \makecell{Motion\\Smooth.$\uparrow$}
& \makecell{Aesthetic\\Quality$\uparrow$} \\
\midrule

\rowcolor{gray!12}
LongLive-5B$^{*}$                   
& 0.5937 & 0.688 & 0.808 & 0.712 & 0.9640 & 0.9731 & 0.9842 & 0.5193 \\
+ B-MST    
& 0.8036 & 0.832 & 0.926 & 0.847 & 0.9694 & 0.9758 & 0.9864 & 0.5281 \\

\bottomrule
\end{tabular}
}
\end{table*}

\subsection{Ablation Studies and Discussion}

\subsubsection{Validation of bias-guided memory-specialized training}
\label{sec:discussion-bmst}
We first validate the effect of bias-guided memory-specialized training (B-MST). As shown in Table~\ref{tab:main_ablation}, all variants are evaluated under the same 60-second setting and use the same total memory budget. To verify that the improvement does not come from additional LoRA fine-tuning alone, we further train the uniform-memory baseline for the same 700 steps with the original sink-local layout, denoted as Uniform-Mem FT. This equal-step control brings only marginal changes over the model before B-MST, indicating that the 300-step DMD stage has largely converged under the uniform-memory design. In contrast, the full model substantially improves both HOI-specific and general metrics. These results show that the gains mainly arise from bias-guided memory specialization under a fixed KV-cache budget, rather than from a larger cache or additional LoRA fine-tuning alone.

\begin{figure}[tb]
  \centering
  \includegraphics[width=\columnwidth, alt={Placeholder main ablation.}]{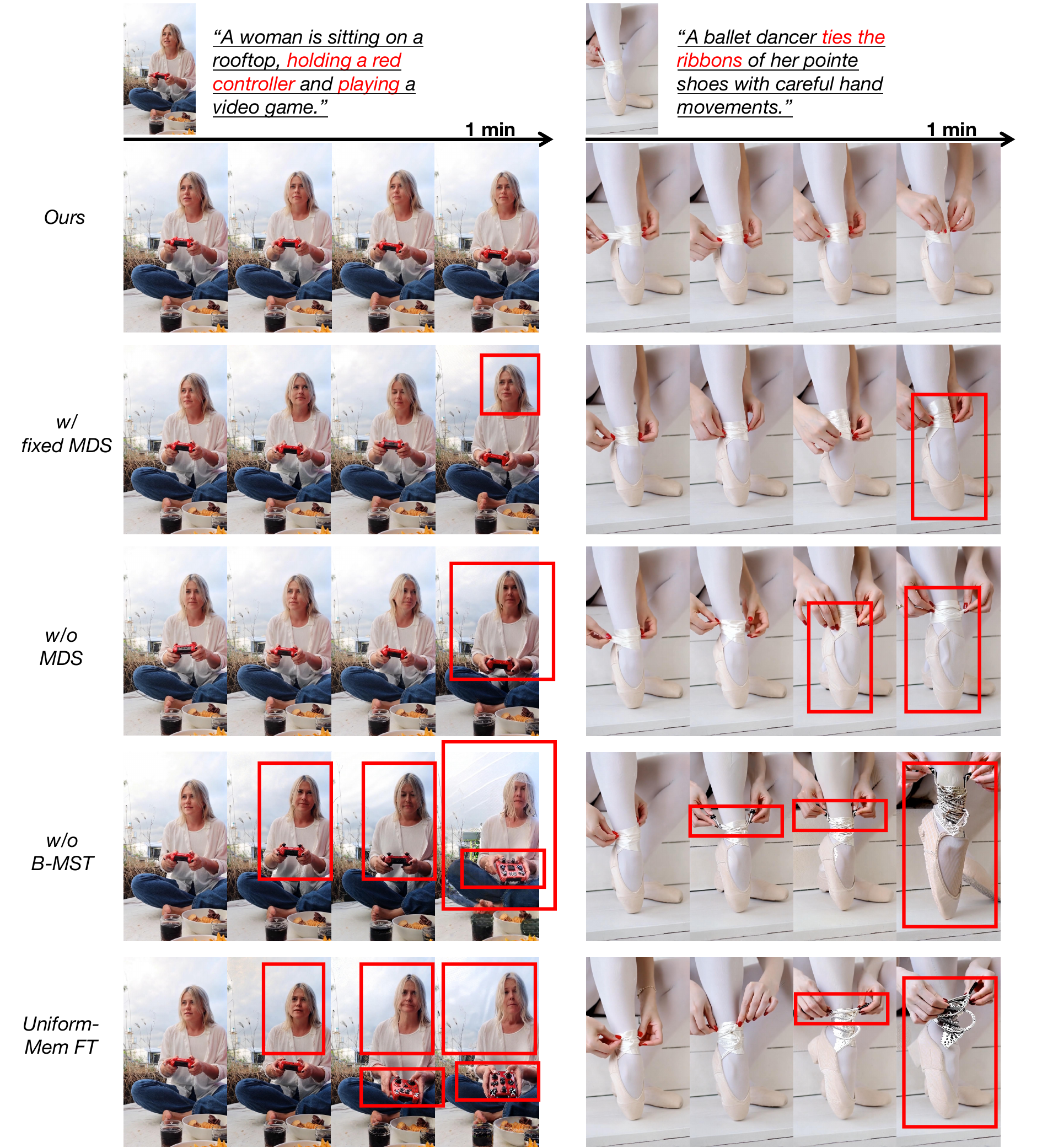}
  \caption{Visual ablation of B-MST and MDS variants. Removing these components leads to interaction collapse and identity drift, whereas our full method maintains robust long-term interactions.}
  \label{fig:main-ablation}
\end{figure}

The qualitative results in Fig.~\ref{fig:main-ablation} are consistent with the quantitative findings. Without B-MST, the model shows attenuation of early interaction priors during long autoregressive generation, leading to unstable hand-object contact, changed object states, and drifted human identity. In contrast, B-MST aligns each Transformer block with its profiled spatial bias: HOI-biased blocks preserve reliable early interaction cues, while surrounding-biased blocks retain recent context for local visual changes and motion continuity. This block-specific memory behavior helps maintain long-term interaction consistency without sacrificing short-term temporal smoothness.

\begin{figure}[tb]
  \centering
  \includegraphics[width=\columnwidth, alt={Placeholder group ablation.}]{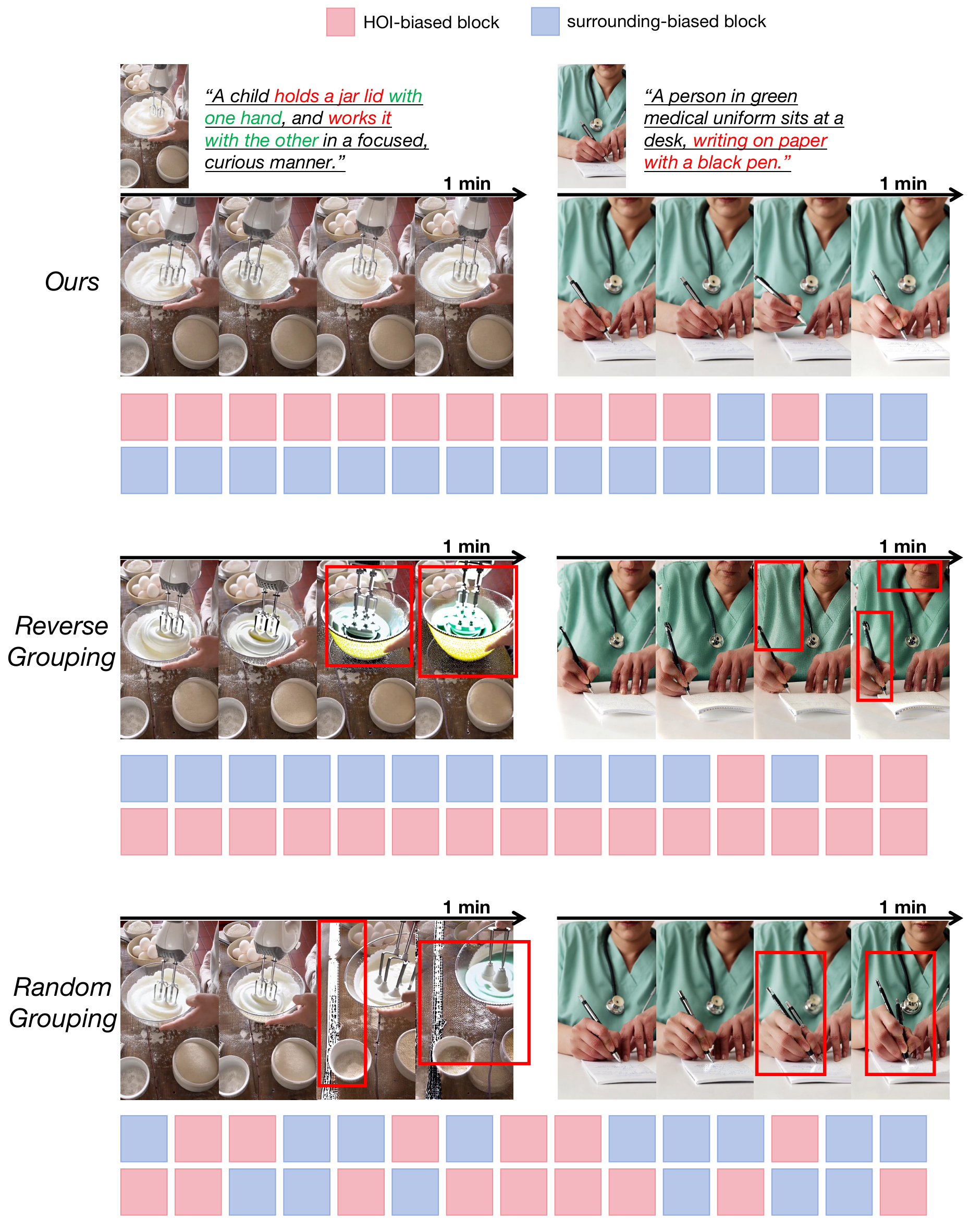}
  \caption{Ablation of block grouping strategies. Mismatching memory allocation with spatial bias (Reverse/Random) causes contact drift, whereas our semantic grouping preserves stable interactions.}
  \label{fig:group-ablation}
\end{figure}

\noindent\textbf{Block grouping verification.}
We further examine whether the proposed memory specialization depends on the semantic grouping of blocks. We compare our semantic grouping with two controls: reverse grouping and random grouping. Reverse grouping exchanges the memory assignments of HOI-biased and surrounding-biased blocks, while random grouping keeps the same group sizes but samples block identities randomly. As reported in Table~\ref{tab:main_ablation}, both controls perform worse than our semantic grouping, with reverse grouping causing the largest degradation in long-term HOI consistency. This indicates that non-uniform memory allocation alone is insufficient; the memory behavior needs to match the semantic bias of each block group. The qualitative results in Fig.~\ref{fig:group-ablation} show the same trend: reverse and random grouping often lead to contact drift, object state changes, or unstable temporal continuity, whereas our grouping better preserves both interaction states and short-term visual consistency.

\begin{figure}[tb]
  \centering
  \includegraphics[width=\columnwidth, alt={Placeholder BMR generalization validation.}]{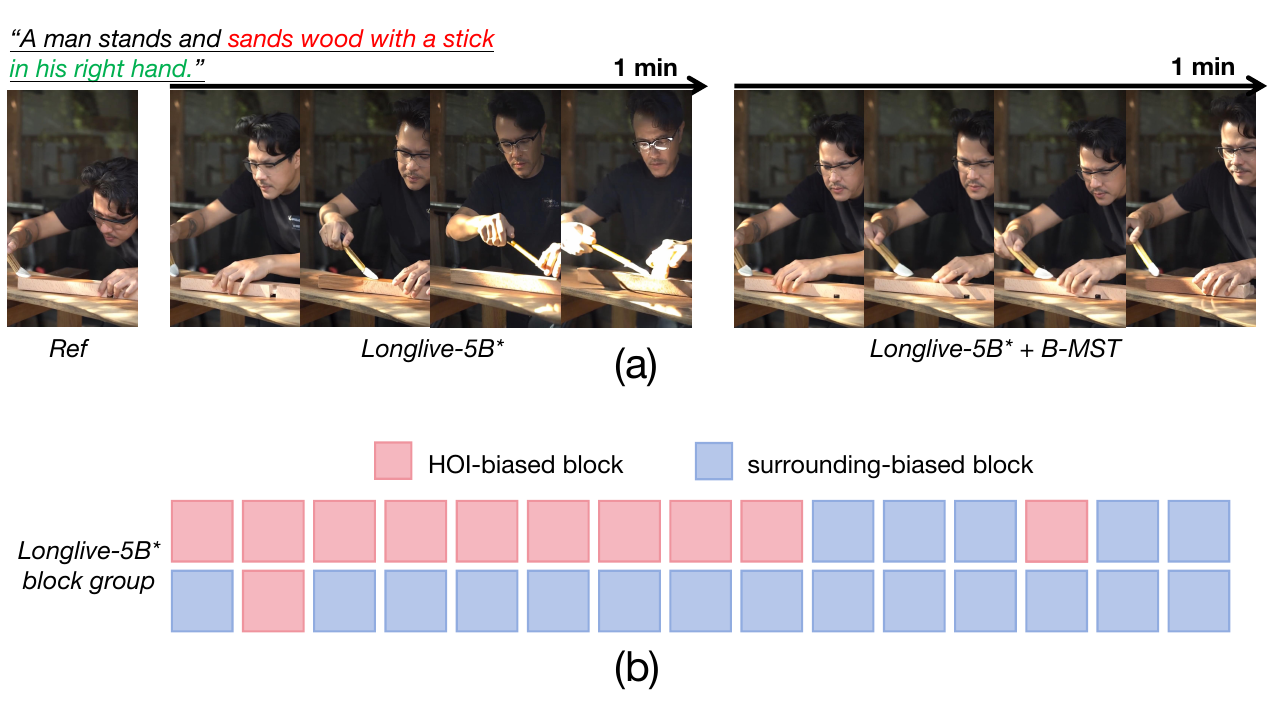}
  \caption{Generalization of B-MST to LongLive-5B*. (a) Integrating B-MST prevents the interaction drift and object distortion observed in the original LongLive-5B* during long-horizon generation. (b) The profiled block grouping for LongLive-5B*.}
  \label{fig:add-longlive}
\end{figure}

\noindent\textbf{Generalization validation.}
To verify that the proposed memory specialization is not tied to a specific backbone, we further apply B-MST to LongLive-5B$^{*}$. 
Specifically, we first apply the same offline HOI-aware block profiling strategy in Sec.~\ref{sec:offline-block-profiling} to LongLive-5B$^{*}$, which identifies its HOI-biased and surrounding-biased blocks. 
The resulting block grouping is shown in Fig.~\ref{fig:add-longlive}(b), and B-MST is then instantiated according to these profiled block types. 
As reported in Table~\ref{tab:add_cache_allocation_to_longlive}, B-MST consistently improves LongLive-5B$^{*}$ across both HOI-specific and general metrics. 
The gains on Obj-CLIP and InternVL-based metrics indicate better preservation of object appearance, human states, and interaction relations, while the improvements on general metrics show more stable foreground and background consistency. 
The qualitative comparison in Fig.~\ref{fig:add-longlive}(a) further shows that the original LongLive-5B$^{*}$ accumulates errors during long generation, leading to unstable hand-object contact and appearance changes. With B-MST, the model better maintains the hand-object interaction and object state across chunks. 
These results show that B-MST can serve as a general memory adaptation strategy for long-term streaming HOI generation.

\begin{figure}[tb]
  \centering
  \includegraphics[width=\columnwidth, alt={Placeholder learnable MDS.}]{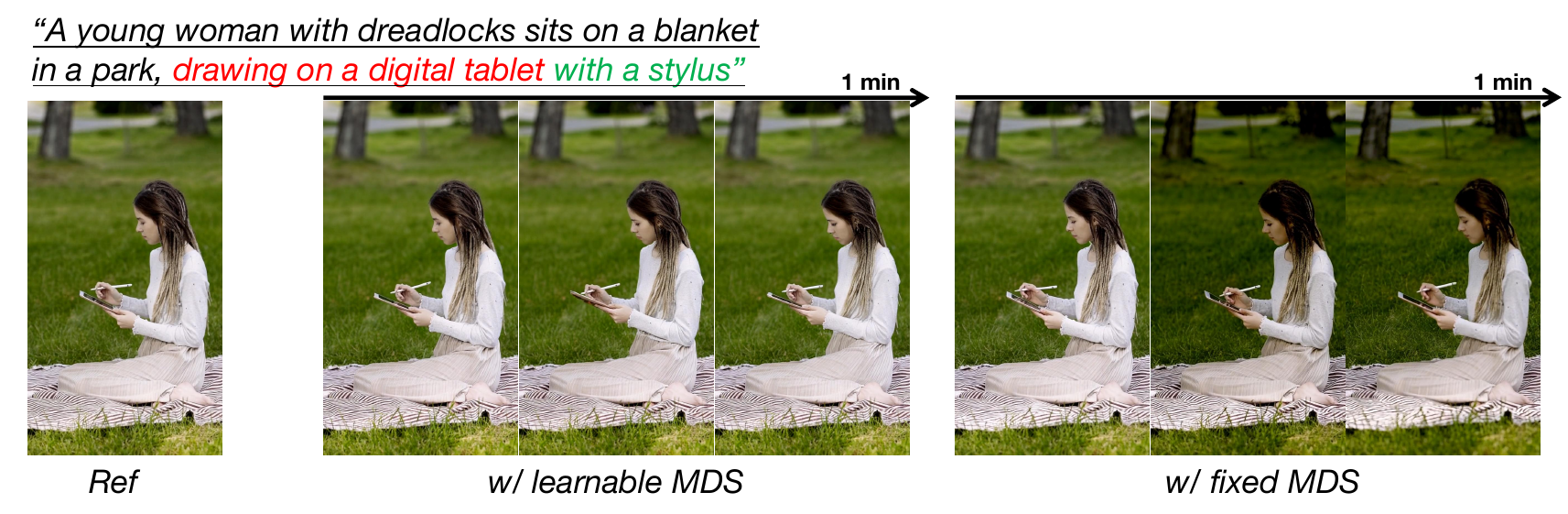}
\caption{Mitigation of temporal degradation using learnable MDS. While the fixed MDS configuration suffers from progressive color drift during extended generation, the learnable MDS mechanism effectively preserves color consistency, resulting in improved long-horizon consistency.}
  \label{fig:learnable}
\end{figure}

\subsubsection{Validation of memory distance scaling}

We then validate the effect of memory distance scaling module (MDS). MDS adjusts the perceived temporal distance between the current chunk and the sink memory of HOI-biased blocks, making early interaction states easier to access during long streaming generation. As shown in Table~\ref{tab:main_ablation}, removing MDS leads to a clear performance drop compared with the full model. This indicates that memory capacity and temporal distance play complementary roles. Block-specific memory allocation provides HOI-biased blocks with more early interaction tokens, while MDS makes these temporally distant tokens more effective during attention.

\noindent\textbf{Effect of learnable MDS.}
We also evaluate a variant with manually fixed MDS parameters. These parameters are set according to the offline profiling results in Sec.~\ref{sec:offline-block-profiling}, with details provided in the supplementary material. As shown in Table~\ref{tab:main_ablation}, the fixed MDS parameter variant already achieves strong performance, suggesting that the profiling results provide useful guidance for configuring memory distance scaling module. However, it is still consistently worse than the learnable MDS used in our full model. This shows that manually designed scaling can provide a reasonable prior, but learning the scaling coefficients allows different HOI-biased blocks to adapt their access to long-range sink memory more effectively.

Fig.~\ref{fig:learnable} highlights a manifestation of temporal degradation in the fixed MDS variant, characterized by progressive color drift over extended generation horizons. After learning the MDS coefficients, our full model better preserves the color appearance and overall visual consistency.

\begin{figure*}[tb]
  \centering
  \includegraphics[width=\textwidth, alt={Placeholder user-study figure.}]{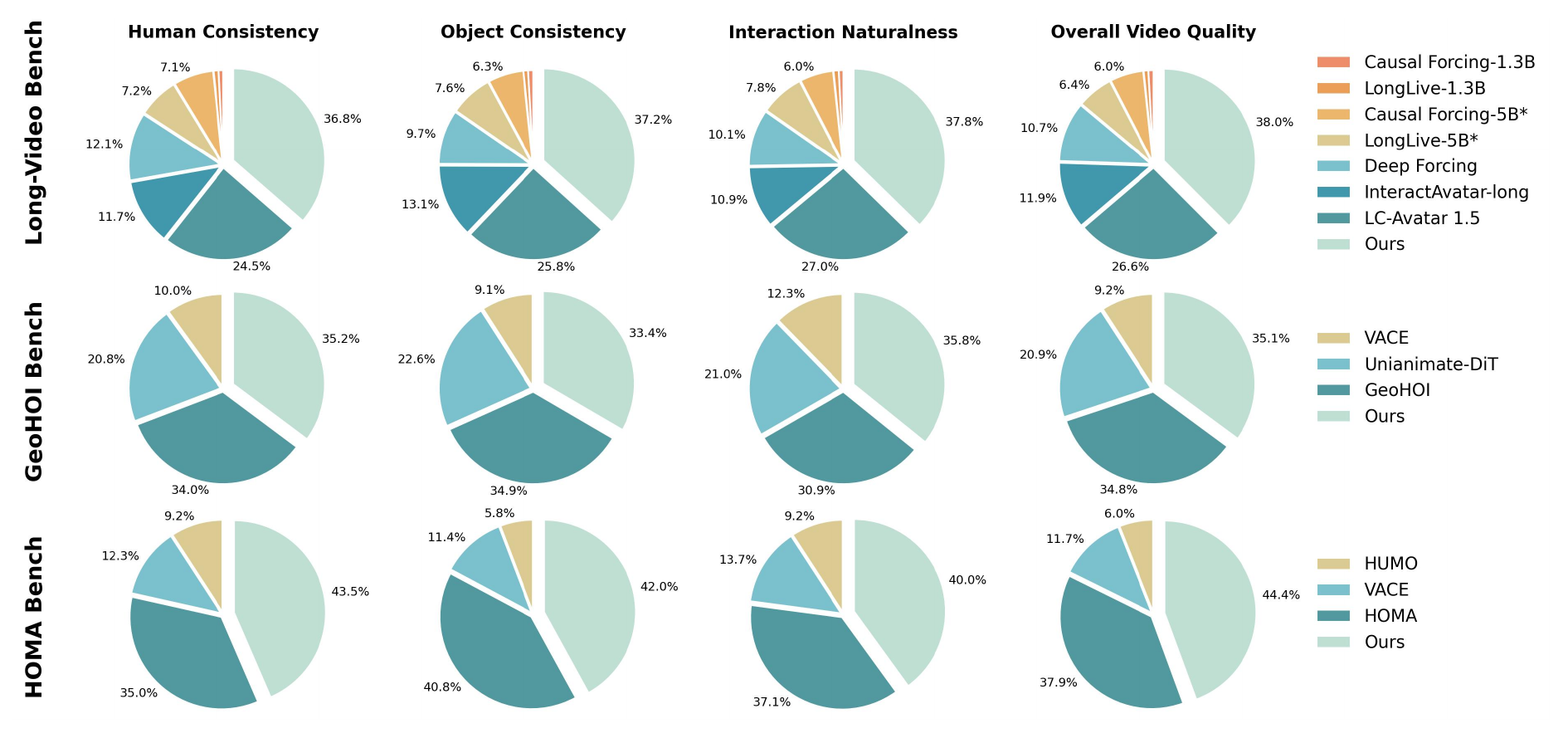}
  \caption{\textbf{User study results.} Pie charts show first-choice preference rates on Long-Video, GeoHOI, and HOMA benchmarks under four perceptual criteria: Human Consistency, Object Consistency, Interaction Naturalness, and Overall Video Quality.}
  \label{fig:user-study}
\end{figure*}

\subsection{User Study}

We conducted a user study to assess the perceptual quality of our method relative to long-video methods, and bidirectional HOI generation methods. We adopted an $N$-way forced-choice protocol. For each image-text pair, participants viewed the videos generated by all competing methods within the corresponding benchmark, presented in randomized order, and selected the single best result according to four criteria: Human Consistency, Object Consistency, Interaction Naturalness, and Overall Video Quality. In total, 28 participants took part in the study, with each participant completing 40 randomized trials across the three benchmarks. Short breaks were provided at regular intervals to reduce fatigue. For each benchmark and criterion, we report the percentage of trials in which each method was selected as the best, referred to as the first-choice preference rate. The results are summarized in Fig.~\ref{fig:user-study}.

The user preferences in Fig.~\ref{fig:user-study} show a consistent advantage for our method on long-video generation. On the \textbf{Long-Video Bench}, our method receives the highest first-choice preference rate across all four criteria, ranging from 36.8\% to 38.0\%. This is substantially higher than the strongest baseline, LC-Avatar 1.5 (24.5\%--27.0\%), and well above the remaining streaming methods. In particular, Causal Forcing and LongLive at the 1.3B scale are almost never selected as the best result ($<$0.4\%), while their 5B variants and Deep Forcing receive only 6\%--13\% of the votes. These results indicate that the proposed memory design more reliably preserves human appearance, object identity, and hand-object interaction states during long video generation.

Our method also remains competitive against dedicated bidirectional HOI generation models. On the \textbf{GeoHOI Bench}, StreamHOI obtains 33.4\%--35.8\% first-choice preference across the four criteria, matching the dedicated GeoHOI model (30.9\%--34.9\%) and outperforming Unianimate-DiT (20.8\%--22.6\%) and VACE (9.1\%--12.3\%). On the \textbf{HOMA Bench}, StreamHOI achieves the highest preference rate for every criterion (40.0\%--44.4\%), ahead of HOMA (35.0\%--40.8\%), whereas VACE and HUMO are selected much less frequently. Overall, the study shows that StreamHOI delivers perceptual quality comparable to, and often preferred over, specialized bidirectional HOI generation methods, while preserving the efficiency and temporal scalability of streaming generation.

\begin{figure}[tb]
  \centering
  \includegraphics[width=\columnwidth, alt={Placeholder Long horizon.}]{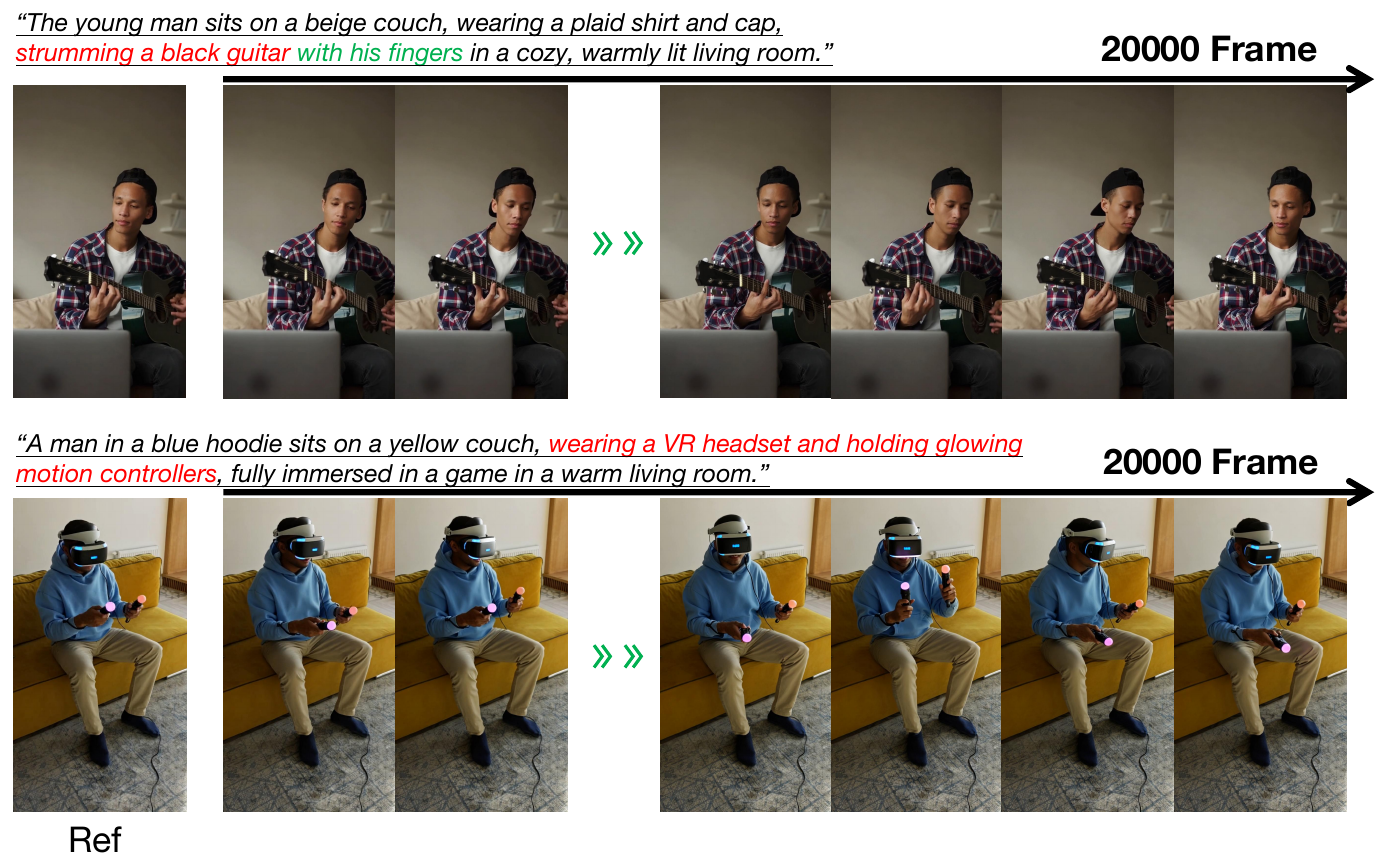}
  \caption{Qualitative results of extended-horizon generation. StreamHOI maintains stable human–object interactions and structural consistency over a 20,000-frame sequence, showing reduced visual degradation during long autoregressive generation.}
  \label{fig:long-horizon}
\end{figure}

\begin{figure}[tb]
  \centering
  \includegraphics[width=\columnwidth, alt={Placeholder Switch prompt.}]{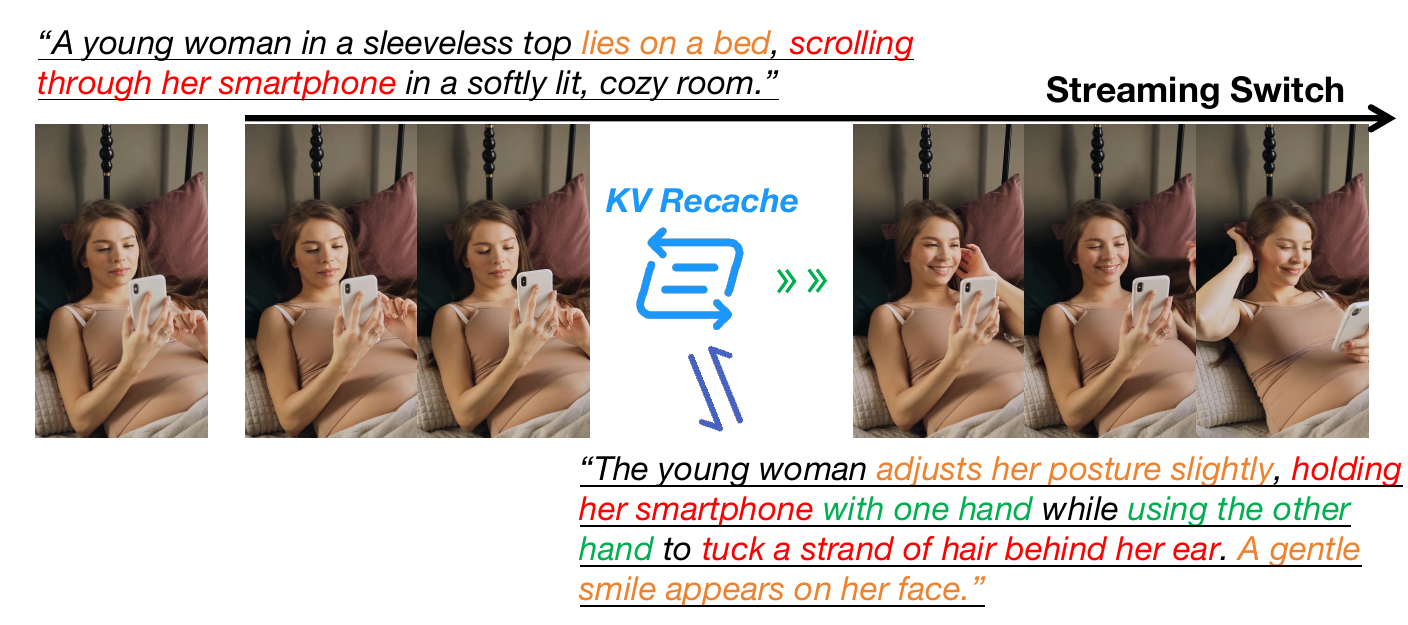}
  \caption{Dynamic prompt switching during streaming generation. Upon receiving a new textual prompt, the model adapts the generated motion and interaction in real time with high interaction quality.}
  \label{fig:switch-prompt}
\end{figure}

\subsection{Long-Horizon Evaluation}

We further evaluate the long-horizon generation ability of our method. As shown in Fig.~\ref{fig:long-horizon}, our model can generate a 20,000-frame video while maintaining overall visual and interaction consistency. During the whole generation process, the model maintains stable human--object interaction quality, consistent object states, and smooth motion transitions. This result shows that the proposed temporal memory adaptation mechanism can effectively support long-duration streaming generation and reduce error accumulation over a very long horizon. Notably, our method can be combined with the KV Recache~\cite{yang2025longlive} mechanism to dynamically switch prompts during streaming generation, enabling streaming control of human--object interactions. As illustrated in Fig.~\ref{fig:switch-prompt}, the model responds immediately upon receiving the next prompt, accurately following the instructed interaction changes.

\subsection{Runtime Analysis}
\label{sec:runtime}
We report the runtime comparison in Table~\ref{tab:fps-comparison}. StreamHOI achieves 17.6 FPS with a first-chunk latency of 0.75s. At the same 5B model scale, it matches the throughput of LongLive$^{*}$ and is faster than Causal Forcing$^{*}$ and Deep Forcing, while providing stronger long-duration HOI quality as shown in the previous comparisons. Although the 1.3B LongLive baseline reaches higher FPS, its generation quality is substantially weaker, especially under long temporal settings. Compared with larger avatar or bidirectional HOI generation methods, StreamHOI is significantly more efficient: LC-Avatar 1.5 requires 87.51s and InteractAvatar-Long requires 600s, while other bidirectional HOI methods require 87s to 549s for end-to-end generation. These results show that StreamHOI preserves the low-latency advantage of streaming generation while achieving competitive long-duration HOI quality.
\begin{table}[tb]
\centering
\caption{FPS and latency comparison of different methods.}
\label{tab:fps-comparison}
\begin{tabular}{lccc}
\toprule
\textbf{Method} & \textbf{Model Size} & \textbf{FPS$\uparrow$} & \textbf{Latency$\downarrow$} \\
\midrule
Causal Forcing & 1.3B & 8.9 & 0.45 \\
LongLive & 1.3B & 20.7 & 0.69 \\
Causal Forcing$^{*}$ & 5B & 14.4 & 0.75 \\
LongLive$^{*}$ & 5B & 17.6 & 0.75 \\
Deep Forcing & 5B & 14.3 & 0.75 \\
LC-Avatar 1.5 & 13.6B & 0.9 & 87.51 \\
InteractAvatar-Long & 10B & 0.14 & 600 \\
\midrule
UniAnimate-DiT & 14B & 0.15 & 549 \\
VACE & 14B & 0.24 & 334 \\
HUMO & 17B & 0.17 & 466 \\
GeoHOI & 5B & 0.93 & 87 \\
HOMA & 13B & 0.25 & 329 \\
\midrule
\rowcolor{orange!12}
StreamHOI & 5B & 17.6 & 0.75 \\
\bottomrule
\end{tabular}
\end{table}

\section{Limitations}
When the human--object interaction in the reference image is unclear, such as ambiguous contact regions or occluded objects, StreamHOI may generate less stable interactions. In addition, since our training data mainly contains single-person videos, StreamHOI may be less effective in complex multi-person scenarios. We leave these cases for future work.
\section{Conclusion}
We present StreamHOI, a streaming framework for long-duration human--object interaction video generation. By analyzing block-wise spatial attention, we showed that different Transformer blocks have distinct memory preferences for HOI region and surrounding region. Based on this observation, StreamHOI uses bias-guided memory-specialized training to adapt the generator to a block-specific sink-local memory layout, and applies memory distance scaling module to strengthen long-range access to reliable interaction anchors. Experiments on long-video and HOI benchmarks show that StreamHOI improves interaction plausibility, object preservation, temporal stability, and runtime efficiency under bounded memory and low-latency generation. These results suggest that interaction-aware temporal memory adaptation is an effective direction for scalable streaming HOI video generation.

{
    \small
    \bibliographystyle{IEEEtran}
    \bibliography{main}
}

\end{document}